\documentclass{article} 
\usepackage[final]{colm2025_conference}

\usepackage{microtype}
\usepackage{hyperref}
\usepackage{url}
\usepackage{booktabs}

\usepackage{lineno}
\usepackage{graphicx}
\usepackage{amsmath}
\usepackage{makecell}
\usepackage{multirow}
\usepackage{amssymb}
\usepackage{colortbl}   
\usepackage{pifont}
\usepackage{enumitem}

\definecolor{darkblue}{rgb}{0, 0, 0.5}
\definecolor{Gray}{gray}{0.93}
\definecolor{uclagold}{rgb}{1.0, 0.7, 0.0}
\definecolor{airforceblue}{rgb}{0.36, 0.54, 0.66}
\definecolor{rosegold}{rgb}{0.72, 0.43, 0.47}
\definecolor{pastelbrown}{rgb}{0.51, 0.41, 0.33}
\definecolor{isabelline}{rgb}{0.96, 0.94, 0.93}
\definecolor{macaroniandcheese}{rgb}{0.98, 0.89, 0.83}
\definecolor{wildblueyonder}{rgb}{0.85, 0.89, 0.95}
\definecolor{mediumtaupe}{rgb}{0.4, 0.3, 0.28}
\definecolor{bluegray}{rgb}{0.4, 0.6, 0.8}
\definecolor{celestialblue}{rgb}{0.29, 0.59, 0.82}
\definecolor{darkorange}{rgb}{1.0, 0.55, 0.0}
\definecolor{cadmiumred}{rgb}{0.89, 0.0, 0.13}
\definecolor{magnolia}{rgb}{0.97, 0.96, 1.0}
\definecolor{pastelblue}{rgb}{0.68, 0.78, 0.81}
\definecolor{persiangreen}{rgb}{0.0, 0.65, 0.58}
\definecolor{steelblue}{rgb}{0.27, 0.51, 0.71}
\definecolor{bluebell}{rgb}{0.64, 0.64, 0.82}
\definecolor{dimgray}{rgb}{0.41, 0.41, 0.41}
\definecolor{splashedwhite}{rgb}{1.0, 0.99, 1.0}
\definecolor{lavendergray}{rgb}{0.77, 0.76, 0.82}
\definecolor{lightgray}{rgb}{0.83, 0.83, 0.83}
\definecolor{lavendermist}{rgb}{0.9, 0.9, 0.98}
\definecolor{lightgreen}{HTML}{f8fcf4}
\definecolor{lightblue}{HTML}{dfebf7}
\hypersetup{colorlinks=true, citecolor=darkblue, linkcolor=darkblue, urlcolor=darkblue}
\title{SlowFast-LLaVA-1.5: A Family of Token-Efficient Video Large Language Models for Long-Form Video Understanding}

\author{Mingze Xu$^\circ$, \,Mingfei Gao$^\circ$,\,Shiyu Li$^\circ$, \,Jiasen Lu,\,\,Zhe Gan,\,Zhengfeng Lai, \\
\textbf{Meng Cao,\, \,Kai Kang, \,Yinfei Yang$^\dagger$, \,Afshin Dehghan$^\dagger$} \\
Apple \\
\small{\texttt{\{mingze\_xu2,mgao22,shiyu\_li,yinfeiy,adehghan\}@apple.com}} \\
$^\circ$First authors; $^\dagger$Senior authors
}

\begin{document}

\ifcolmsubmission
\linenumbers
\fi

\maketitle

\begin{abstract}
We introduce \textbf{SlowFast-LLaVA-1.5} (abbreviated as SF-LLaVA-1.5), a family of video large language models (LLMs) offering a token-efficient solution for long-form video understanding. We incorporate the two-stream SlowFast mechanism into a streamlined training pipeline, and perform joint video-image training on a carefully curated data mixture of only publicly available datasets. Our primary focus is on highly efficient model scales (1B and 3B), demonstrating that even relatively small Video LLMs can achieve state-of-the-art performance on video understanding, meeting the demand for mobile-friendly models. Experimental results demonstrate that SF-LLaVA-1.5 achieves superior performance on a wide range of video and image tasks, with robust results at all model sizes (ranging from 1B to 7B). Notably, SF-LLaVA-1.5 achieves state-of-the-art results in long-form video understanding (\textit{e.g.,} LongVideoBench and MLVU)
and excels at small scales across various video benchmarks.
\end{abstract}

\vspace{-7pt}
\section{Introduction}
\label{sec:introduction}
\vspace{-5pt}

Video large language models (LLMs)~\citep{Maaz2023VideoChatGPT,lin2023video,xu2024pllava}
integrate video perception into pre-trained LLMs to process videos and generate responses to user commands.
Although significant progress has been made, notable limitations remain in existing Video LLMs.
\textit{First}, they enhance perception and reasoning over long video sequences by leveraging the LLM's increasing context length and handling massive input frames~\citep{shen2024longvu,chen2024longvila,zhang2024long}.
However, the potential for transferring this capability to highly efficient models is underexplored.
\textit{Second}, achieving optimal performance typically requires internal datasets and a complex training lifecycle, with selective parameters frozen at each stage~\citep{li2024llava,zhang2025videollama3}.
These intricate designs lead to high computational costs and reproducibility challenges.
\textit{Third}, many Video LLMs~\citep{zohar2024apollo,li2024videochat} are optimized exclusively for video tasks, limiting their effectiveness as joint models for image understanding tasks.

Building upon the success of SlowFast-LLaVA~\citep{xu2024slowfast}, we introduce \textbf{SlowFast-LLaVA-1.5},
a new family of Video LLMs for long-form video understanding,
focusing on the most efficient model scales (1B and 3B).
Our model family is both effective and token-efficient in modeling long-range temporal context.
This is achieved by employing the SlowFast mechanism, which balances the trade-off between
processing more input frames that significantly increases the token count and computational cost,
and reducing tokens per frame that inevitably loses fine-grained details.
Specifically, the Slow pathway captures detailed spatial features at a low frame rate,
while the Fast pathway operates at a high frame rate with fewer tokens per frame to focus on motion cues.
The success of our model also relies on a streamlined training pipeline
and a carefully curated data mixture.
Our model training consists of only two stages.
The first stage is supervised fine-tuning on image-only data,
providing a good foundation for general knowledge and reasoning.
The second stage conducts video-image joint training
to learn spatial and temporal features for video understanding while maintaining strong performance in image understanding.
To ensure seamless reproducibility,
all pre-trained weights and training datasets used in this work are publicly accessible.

We comprehensively evaluate our models on various video and image benchmarks.
Experimental results demonstrate that SlowFast-LLaVA-1.5 achieves state-of-the-art performance
in long-form video understanding.
Notably, our 7B model scores 62.5\% on LongVideoBench and 71.5\% on MLVU,
outperforming existing methods by a clear margin.
SlowFast-LLaVA-1.5 also excels at smaller model sizes,
achieving 56.6\% and 60.8\% on Video-MME (w/o sub) at the 1B and 3B scales, respectively.
As a unified image and video model, it maintains strong image performance despite the simple training recipe.

Our main contributions are as follows.
\textit{First}, we introduce SlowFast-LLaVA-1.5, a new family of Video LLMs ranging from 1B to 7B parameters.
We demonstrate the effectiveness of incorporating the SlowFast mechanism into a supervised fine-tuning framework,
modeling long-range context while maintaining high efficiency.
\textit{Second}, our model family provides enhanced reproducibility by using only two training stages and publicly available datasets, distinguishing it from existing methods.
\textit{Third}, SlowFast-LLaVA-1.5 achieves the state-of-the-art performance on long-form video understanding. Moreover, our smaller models (1B and 3B) clearly outperform comparable Video LLMs across video benchmarks.

\vspace{-7pt}
\section{Related Work}
\label{sec:related_work}
\vspace{-9pt}

\textbf{Image Large Language Models} have gained widespread attention~\citep{achiam2023gpt,team2023gemini,touvron2023llama,chen2024expanding,bai2025qwen2}.
Significant progress across multiple fronts includes:
\textit{(i)} enhancing data quantity and quality during pre-training~\citep{mckinzie2024mm1,liu2024llavanext,lin2023vila,li2024omnicorpus}
and supervised fine-tuning (SFT)~\citep{zhang2024mm1,deitke2024molmo,chen2024sharegpt4v,wang2023instruct4v,tong2025cambrian};
\textit{(ii)} accommodating images of various high resolutions~\citep{lin2023sphinx,zhang2024internlm,wang2024qwen2};
\textit{(iii)} improving architecture designs, including different visual encoders~\citep{zhai2023sigmoid,tong2024eyes,shi2024eagle} and vision-language connectors~\citep{li2023blip2,cha2024honeybee};
and \textit{iv)} conducting comprehensive studies for easy deployment~\citep{team2023gemini,marafioti2025smolvlm}.
These rapid advancements also establish a strong foundation for related areas
such as video understanding~\citep{Maaz2023VideoChatGPT,lin2023video}, referring \& grounding~\citep{you2023ferret,you2024ferret}, and visual agents~\citep{durante2024agent,yang2025magma}.

\textbf{Video Large Language Models} have become an active research area~\citep{2023videochat,song2024moviechat,chen2024sharegpt4video,zhang2024direct,zohar2024apollo}.
Early Video LLMs are developed as specialist models~\citep{damonlpsg2023videollama,damonlpsg2024videollama2,xu2024pllava,ryoo2024xgen},
achieving strong performance on video tasks but with some trade-offs in image understanding.
Training-free Video LLMs~\citep{kim2024image,xu2024slowfast} offer an efficient alternative by leveraging Image LLMs without fine-tuning on video data, enabling flexible deployment across various applications.
Recent models~\citep{llava178k,liu2024oryx,zhang2025videollama3} are jointly trained on video and image datasets, obtaining superior results in both modalities.
Long-form video understanding~\citep{zhou2024mlvu,wu2025longvideobench} gained increasing attentions, addressing hour-long videos~\citep{chen2024longvila,li2024videochat} or live streams~\citep{qian2024streaming,zhang2024flash} while optimizing the token efficiency~\citep{lee2024video}.
The proposed SlowFast-LLaVA-1.5 is a family of Video LLMs designed for modeling long-range temporal context.
It enhances SlowFast-LLaVA~\citep{xu2024slowfast} by implementing the SlowFast design within a unified video-image training framework, achieving state-of-the-art performance with efficient token utilization.

\vspace{-7pt}
\section{\textbf{S}low\textbf{F}ast-LLaVA-1.5}
\label{sec:sf_llava}
\vspace{-9pt}

We provide a detailed explanation of \textbf{S}low\textbf{F}ast-LLaVA-1.5 (abbreviated as SF-LLaVA-1.5), which incorporates the SlowFast video projector into a LLaVA-style architecture, improving long-range temporal modeling while optimizing token efficiency.
In contrast to its training-free pioneer~\citep{xu2024slowfast}, this paper \textit{(i)} systematically investigates different instantiations based on the generic SlowFast idea (Sec.~\ref{sec:model_arch}), \textit{(ii)} designs a compact yet effective training pipeline (Sec.~\ref{sec:training_pipeline}),
and \textit{(iii)} introduces tailored data mixtures using only publicly available datasets for each training stage (Sec.~\ref{sec:data_mixture}).

\begin{figure*}[t]
    \centering
    \includegraphics[width=1.0\textwidth]{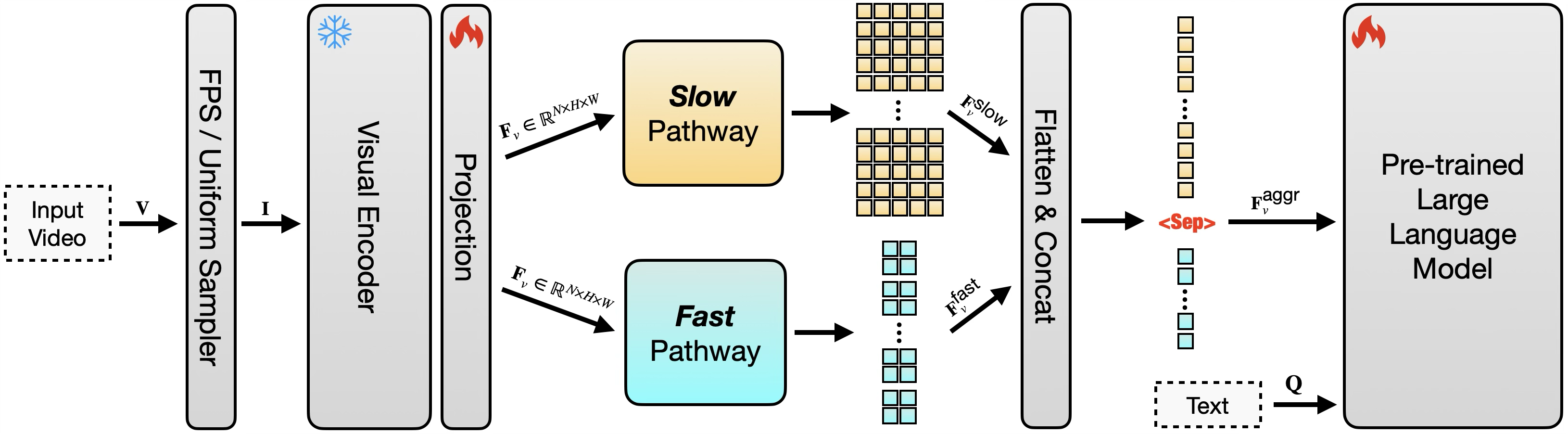}
    \vspace{-15pt}
    \caption{
        \textbf{Visualization of the video understanding pipeline in SlowFast-LLaVA-1.5.}~Compared to its training-free pioneer~\citep{xu2024slowfast}, our projector and LLM are fine-tuned throughout the training cycle, while keeping the vision encoder frozen.
    }
    \label{fig:overview}
    \vspace{-5pt}
\end{figure*}

\vspace{-5pt}
\subsection{Model Architecture}
\label{sec:model_arch}
\vspace{-5pt}

As shown in Fig.~\ref{fig:overview}, the architecture of SF-LLaVA-1.5 follows the core design principle of SF-LLaVA~\citep{xu2024slowfast}.
It takes a video/image $\mathbf{V}$ and a question $\mathbf{Q}$ as inputs and responds with a textual answer $\mathbf{A}$.
For video inputs, we sample $N$ frames, $\mathbf{I} = \{I_1, I_2, ..., I_N\}$, at a fixed frame rate without special frame assembling ($N$ equals $1$ for image input).
After that, a visual encoder (\textit{e.g.,} OryxViT~\citep{liu2024oryx}) is used to extract frame-level features $\mathbf{F}_v \in \mathbb{R}^{N\times H \times W}$ from the inputs independently, keeping their original aspect ratio.
The video and image feature tokens are then fed into different projectors, with video using the two-stream SlowFast projector and image using a two-layer MLP.

\textbf{The SlowFast projector} processes $\mathbf{F}_v$ through two pathways, one dedicated to capturing spatial patterns and the other to modeling motion cues.
\begin{itemize}[leftmargin=22pt]
    \item \textbf{The Slow pathway}, which focuses on capturing detailed spatial semantics, operates at a reduced frame rate by downsampling the total frame count from $N$ to $N^{slow}$. To further improve the efficiency while preserving sufficient details, it applies spatial pooling over $\mathbf{F}_{v}$ with proper strides of $\sigma_h \times \sigma_w$. The output feature is $\mathbf{F}_{v}^{\textnormal{slow}} \in \mathbb{R}^{N^{\textnormal{slow}} \times H^{\textnormal{slow}} \times W^{\textnormal{slow}}}$, where $H^{\textnormal{slow}} = H / \sigma_h$ and $W^{\textnormal{slow}} = W / \sigma_w$.
    \item \textbf{The Fast Pathway}, which focuses on modeling long-range context, maintains the original frame rate, while downsampling more aggressively on the spatial resolution to $H^{\textnormal{fast}}\times W^{\textnormal{fast}} $. The output feature is $\mathbf{F}_{v}^{\textnormal{fast}} \in \mathbb{R}^{N^{\textnormal{fast}} \times H^{\textnormal{fast}} \times W^{\textnormal{fast}}}$, where $N^{\textnormal{fast}} = N$, $H^{\textnormal{fast}} \ll H^{\textnormal{slow}}$ and $W^{\textnormal{fast}} \ll W^{\textnormal{slow}}$.
\end{itemize}
$\mathbf{F}_{v}^{\textnormal{slow}}$ and $\mathbf{F}_{v}^{\textnormal{fast}}$ are flattened and concatenated together as a token vector $\mathbf{F}_{v}^{\textnormal{aggr}}$, which serves as the final visual input to the LLM. A dedicated special token is typically used to separate $\mathbf{F}_{v}^{\textnormal{slow}}$ and $\mathbf{F}_{v}^{\textnormal{fast}}$, assisting the LLM in distinguishing the two sets of features.

\vspace{-5pt}
\subsubsection{Instantiations of SlowFast}
\label{sec:slowfast_instantiations}
\vspace{-2pt}

Next, we describe two approaches for organizing the Slow and Fast tokens.

\begin{itemize}[leftmargin=22pt]
    \item \textbf{The Group-based SlowFast (GSF)} places the Slow tokens before the Fast tokens (Appendix Fig.~\ref{fig:gsf_vs_isf} above). This design is inspired by the AnyRes~\citep{zhang2024llavanextvideo} technique in image understanding, where the Fast tokens provide a global overview of the video and the Slow tokens capture fine-grained spatial details. Notably, SlowFast-LLaVA~\citep{xu2024slowfast} works effectively only under this setting, as it is a training-free model that benefits from ``overfitting'' to its image backbone (\textit{i.e.,} LLaVA-NeXT).
    \item \textbf{The Interleaved SlowFast (ISF)} arranges the tokens according to their spatial and temporal order (Appendix Fig.~\ref{fig:gsf_vs_isf} below). Since Slow and Fast frames contain different numbers of tokens, a learnable special token is utilized to separate adjacent frames, allowing the LLM to distinguish which frame a token belongs to. Different from GSF, $N^{fast}$ equals to $N - N^{slow}$ in this approach. ISF balances the presence of both token types throughout the input sequence, preventing the model from becoming overly focused on one type of information at a time.
\end{itemize}

Unless noted otherwise, we use GSF by default,
as it aligns better with the image pipeline using AnyRes inputs.
Interestingly, experiments (Sec.~\ref{sec:slowfast_design_choices}) show that
SF-LLaVA-1.5 is not sensitive to this setting,
suggesting that the generic SlowFast idea and our training recipe are
the main reason for the strong performance on long-form video understanding.

\vspace{-5pt}
\subsection{Training Pipeline}
\label{sec:training_pipeline}
\vspace{-5pt}

The training pipeline of SF-LLaVA-1.5 is much simpler than most of the existing Video LLMs~\citep{chen2024expanding,zohar2024apollo,zhang2025videollama3,li2024videochat}
with only two training stages, as detailed in Table~\ref{tab:training_settings}.

\begin{table}[h]
    \footnotesize
    \centering
    \begin{tabular}{l|cc}
        \hline
        \textbf{Settings} & \textbf{Stage I} & \textbf{Stage II} \\
        \hline
        \textbf{Dataset} & Image & Image \& Video \\
        \textbf{Trainable} & Projector \& LLM & Projector \& LLM\\
        \textbf{Image Projector} & MLP w/ GELU & MLP w/ GELU \\
        \textbf{Video Projector} & - & \makecell{SlowFast} \\
        \textbf{Batch Size} & 512 & 512 \\
        \textbf{Learning Rate} & $2e^{-5}$ & $2e^{-5}$ \\
        \textbf{Context Length} & 8K & 16K \\
        \textbf{Number of Input Frames} & 1 & 1 or 128 \\
        \textbf{Max Image Resolution} & $1280\times1280$& $1536\times1536$ \\
        \textbf{Max Video Resolution} & - &  $480\times480$ \\
        \textbf{Training Steps} & 1 epoch & 1 epoch \\
        \hline
    \end{tabular}
    \vspace{-7pt}
    \caption{\textbf{Training settings for SlowFast-LLaVA-1.5.}}
    \label{tab:training_settings}
\end{table}

\textbf{Stage I (image understanding)} conducts SFT with images to provide a good warmup status for video understanding. For simplicity and efficiency, we do not use any extra pretraining stages~\citep{li2024llava} or image splitting strategy~\citep{lin2023sphinx},
although they have proven to be effective for boosting text-rich results.
Instead, we use native resolution inputs following Oryx~\citep{liu2024oryx},
where, for each image $I_i\in\mathbb{R}^{H_i\times W_i}$, we have a low resolution $I_i^{low}$ and a high resolution $I_i^{high}$ input.
The low-resolution image is obtained by simply resizing the original image to a base resolution, as in $I_i^{low}=\textnormal{resize}(I_i, H_i^{base}\times H_i^{base})$.
For $I_i^{high}$, we keep its original aspect ratio and resize it to $H_i^{high}\times W_i^{high}$, as in Eq.~\ref{eq:scale} and~\ref{eq:resize},
\begin{equation}
\small
    \label{eq:scale}
    scale=
    \begin{cases}
        \sqrt{\theta^{I^{max}}/(H_i\times W_i)}, & \text{if } H_i\times W_i > \theta^{I^{max}}\\
        \sqrt{\theta^{I^{min}}/(H_i\times W_i)}, & \text{if } H_i\times W_i < \theta^{I^{min}}\\
        1.0, & \text{otherwise},
    \end{cases}
\end{equation}
\begin{equation}
\small
    \label{eq:resize}
    \begin{split}
        H_i^{high} & = int(H_i * scale / p) * p \\
        W_i^{high} & = int(W_i * scale / p) * p,
    \end{split}
\end{equation}
where $H_i^{high}$ and $W_i^{high}$ represent the resized heights and widths and $p$ denotes the patch size of the ViT-based vision encoder.
Eq.~\ref{eq:scale} calculates the resizing scale ensuring that the area of $I_i^{high}$ is between two pre-defined, minimum area $\theta^{I^{min}}$ and maximum area $\theta^{I^{max}}$ thresholds.
Eq.~\ref{eq:resize} makes sure that both $H_i^{high}$ and $W_i^{high}$ are multiples of $p$. To accommodate different input resolutions, the original position embeddings of the vision encoder are rescaled using bilinear interpolation. After feature projection, the low-resolution and high-resolution image features are concatenated together as the final image feature.

\textbf{Stage II (joint image $\&$ video understanding)} performs SFT jointly with images and videos,
initialized by the pre-trained checkpoint from Stage I.
By default, we keep the image resizing setting the same as Stage I, except that we increase the maximum area threshold, $\theta^{I^{max}}$, to a larger value for better performance.
For video, each frame uses a single resolution that is set using the same strategy as in Eq.~\ref{eq:scale} and~\ref{eq:resize},
where we use $\theta^{V^{min}}$ and $\theta^{V^{max}}$ to denote the corresponding minimum and maximum area thresholds.

\vspace{-2pt}
\subsection{Data Mixture}
\label{sec:data_mixture}
\vspace{-5pt}

Our image and video mixtures are detailed in Table~\ref{tab:data_mixture}.
Many state-of-the-art models~\citep{li2024llava,zhang2025videollama3} achieve superior performance using internal training data that is unavailable to the research community.
\textbf{To ensure the reproducibility of our models, we only include publicly available datasets in our data mixtures.}

\textbf{Image Mixture.}
General, TextRich, and Knowledge are fundamental for developing the reasoning capabilities of a multimodal LLM, which can ultimately benefit both image and video understanding.
We begin with datasets from these three categories in MM1.5~\citep{zhang2024mm1} and evaluate additional datasets for each group from LLaVA-OneVision~\citep{li2024llava} and InternVL2.5~\citep{chen2024expanding}.
Datasets are included in our mixture only if they empirically improve performance.
The final mixture contains 4.67M samples.

\textbf{Video Mixture.}
We build a diverse set of video instruction-following datasets.
We begin with LLaVA-Hound~\citep{zhang2024direct}, ShareGPT4Video~\citep{chen2024sharegpt4video}, VideoChatGPT-Plus~\citep{Maaz2024VideoGPT+}, and ActivityNet-QA~\citep{yu2019activityqa} to include large-scale video data
with caption and QA labels.
We add NExT-QA~\citep{xiao2021next} and Perception Test~\citep{patraucean2023perception}
to improve performance on temporal reasoning.
Furthermore, we incorporate LLaVA-Video-178K~\citep{llava178k} and Cinepile~\citep{rawal2024cinepile} to enhance long-form video understanding.
Finally, we filter out duplicate videos from the same data source and construct our final mixture with 2.01M training samples.

\vspace{-7pt}
\section{Experiments}
\label{sec:experiments}
\vspace{-7pt}

We evaluate SF-LLaVA-1.5 across multiple video and image QA benchmarks (details will be provided in Appendix~\ref{appendix:benchmarks_and_metrics}).
For video, we focus on long-form video understanding,
while also reporting the results in general video QA and temporal reasoning.
For image, we evaluate the models from general, knowledge, and text-rich perspectives.

\vspace{-5pt}
\subsection{Implementation Details}
\label{sec:implementation_details}
\vspace{-7pt}

\textbf{Model Architecture.} We use Oryx-ViT\footnote{ https://huggingface.co/THUdyh/Oryx-ViT.}~\citep{liu2024oryx} with patch size 16 as visual encoder and Qwen2.5\footnote{https://huggingface.co/Qwen.}~\citep{bai2025qwen2} series of LLMs at varying scales as the backbone.
We employ different projectors for video and image inputs.
Specifically, the Group-based SlowFast (GSF) structure is used to aggregate video tokens.
For the Slow pathway, we uniformly select $N^{slow}=32$ frames and apply $2\times2$ pooling to their extracted features.
For the Fast pathway, we use features of all frames (\textit{i.e.,} $N^{fast}=N=128$) and downsample their features to $4\times4$ tokens.
For the image projector, we use a two-layer MLP with GELU activation function.

\textbf{Training Details.}
As summarized in Table~\ref{tab:training_settings}, we freeze the visual encoder in all stages and only fine-tune the projectors and LLM. We use the same hyperparameters for 1B, 3B, and 7B models, setting the total batch size to 512 and learning rate to $2e^{-5}$.
All models are trained on 128 H100-80G GPUs for 1 epoch.

\vspace{-9pt}
\begin{itemize}[leftmargin=22pt]
\item \textbf{Training Stage I}
only uses image understanding data. The low resolution image $I_i^{low}$ is fixed at $H_i^{base}\times W_i^{base}=384\times384$ and the high resolution image $I_i^{high}$ is obtained as in Eq.~\ref{eq:scale} and Eq.~\ref{eq:resize}, where $\theta^{I^{min}}=0$ and $\theta^{I^{max}}=1280^2$. The maximum context length is set to 8K.
The models trained by this stage are named as \textbf{SF-LLaVA-1.5-Image}.
\item \textbf{Training Stage II}
continues training based on SF-LLaVA-1.5-Image by combining our video and image data mixture.
For image, the high-resolution image is obtained in the same way as Stage I,
except that we increase $\theta^{I^{max}}$ to $1536^2$.
For video, we follow prior work~\citep{zohar2024apollo} and sample frames at 1 FPS. We set the max frame number to 128 and uniformly sample the frames if the number exceeds this upper bound.
For each video frame, we set $\theta^{V^{min}}=288^2$ and $\theta^{V^{max}}=480^2$. The maximum context length is set to 16K.
The models trained by this stage are named as \textbf{SF-LLaVA-1.5}.
\end{itemize}

\vspace{-15pt}
\subsection{Video Understanding Results}
\label{sec:video_results}
\vspace{-5pt}

We mainly compare SF-LLaVA-1.5 with state-of-the-art Video LLMs that are trained on publicly available datasets.
Here we highlight some key observations based on Table~\ref{tab:video_results}.

\begingroup
\setlength{\tabcolsep}{4.5pt}
\begin{table}[t]
    \centering
    \resizebox{\linewidth}{!}{%
    \begin{tabular}{l|c|c|cc|ccc|cc}
        \toprule
        \multirow{4}{*}{Model} & \multirow{4}{*}{\makecell{Max\\Input\\Frames}} & \multirow{4}{*}{\makecell{Max\\Input\\Tokens}} & \multicolumn{2}{c|}{General VideoQA} & \multicolumn{3}{c|}{Long-Form Video Understanding} & \multicolumn{2}{c}{Temporal Reasoning} \\
        \cmidrule{4-10}
        & & & \makecell{VideoMME\\(w/o sub)} & \makecell{PercepTest\\(val)} & \makecell{LongVideoBench\\(val)} & \makecell{MLVU\\(m-avg)} & \makecell{LVBench\\(avg)} & \makecell{TempComp\\(mc)} & \makecell{NExT-QA\\(test)} \\
        \midrule
        \multicolumn{10}{c}{\textit{1B Model Comparison}}\\
        \midrule
        LLaVA-OV-0.5B~\citep{li2024llava} & 32 & 6K & 44.0 & 49.2	& 45.8 & 50.3 & \;\,32.7$^{\dagger}$ & 53.2 & 57.2  \\
        MM1.5-1B~\citep{zhang2024mm1} & 24 & 3K & 45.7 & - & 43.9 & - & - & - & 71.8 \\
        LinVT-Mipha-1.6B~\citep{gao2024linvt} & 120 & - & 44.5 & - & 49.7 & 56.2 & - & 45.2 & 71.1 \\
        Apollo-1.5B~\citep{zohar2024apollo} & 2fps & 3K & 53.0 & \underline{61.0} & \underline{54.1} & \underline{63.3} & - & \textbf{60.8} & - \\
        InternVL2.5-2B~\citep{chen2024expanding} & 64 & 16K & 51.9 & - & 52.0 & 61.4 & \;\,37.9$^{\dagger}$ & \;\,53.4$^{\dagger}$ & \;\,\textbf{77.2}$^{\dagger}$ \\
        Qwen2-VL-2B~\citep{wang2024qwen2} & 2fps & 16K & \underline{55.6} & 53.9 & \;\,48.7$^{\dagger}$ & \;\,62.7$^{\dagger}$ & \;\,\underline{39.4}$^{\dagger}$ & \;\,\underline{60.6}$^{\dagger}$ & \;\,\textbf{77.2}$^{\dagger}$ \\
        \rowcolor{lightblue} 
        SF-LLaVA-1.5-1B & 128 & 9K & \textbf{56.6} & \textbf{61.9} & \textbf{54.3} & \textbf{64.3} &  \textbf{39.7} & 60.5 & \underline{76.7} \\
        \midrule
        \multicolumn{10}{c}{\textit{3B Model Comparison}}\\
        \midrule
        VILA1.5-3B~\citep{liu2024nvila} & 8 & 2K & 42.2 & 49.1 & 42.9 & 44.4 & - & 56.1 & - \\
        MM1.5-3B~\citep{zhang2024mm1} & 24 & 3K & 49.5 & - & 45.4 & - & - & - & 74.7 \\
        LongVU-3.2B~\citep{shen2024longvu} & 1fps & 8K & 51.5 & - & - & 55.9 & - & - & - \\
        InternVL2-4B~\citep{chen2024far} & 64 & 16K & 53.9 & \;\,53.9$^{\dagger}$ & 53.0 & 59.9 & \;\,\underline{35.1}$^{\dagger}$ & \;\,60.2$^{\dagger}$ & \;\,71.1$^{\dagger}$ \\
        LinVT-Blip3-4B~\citep{zohar2024apollo} & 120 & - & 58.3 & - & \underline{56.6} & 67.9 & - & 59.6 & \underline{80.1} \\
        Apollo-3B~\citep{zohar2024apollo} & 2fps & 3K & \underline{58.4} & \underline{65.0} & 55.1 & \underline{68.7} & - & \underline{62.5} & -\\
        \rowcolor{lightblue} 
        SF-LLaVA-1.5-3B & 128 & 9K & \textbf{60.8} & \textbf{65.8} & \textbf{57.3} & \textbf{68.8} & \textbf{43.3} & \textbf{64.0} & \textbf{80.8} \\
        \midrule
        \multicolumn{10}{c}{\textit{7B Model Comparison}}\\
        \midrule
        MM1.5-7B~\citep{zhang2024mm1} & 24 & 3K & 53.5 & - & 49.4 & - & - & - & 76.9 \\
        Kangaroo-8B~\citep{liu2024kangaroo} & 64 & 10K & 56.0 & - & 54.8 & 61.0 & 39.4 & 62.5 & - \\
        Oryx1.5-7B~\citep{liu2024oryx} & 64 & 14K & 58.8 & \textbf{70.0} & 56.3 & 67.5 & \;\,39.0$^{\dagger}$ & \;\,58.8$^{\dagger}$ & 81.8 \\
        LLaVA-OV-7B~\citep{li2024llava} & 32 & 6K & 58.2 & 49.7 & 56.5 & 64.7 & - & - & 79.4 \\
        LLaVA-Video-7B~\citep{llava178k} & 64 & 11K & 63.3 & 66.9 &  58.2 & 70.8 & - & - & 83.2 \\
        Apollo-7B~\citep{zohar2024apollo} & 2fps & 2K & 61.3 & 67.3 & 58.5 & \underline{70.9} & - & 64.9 & - \\
        NVILA-8B~\citep{liu2024nvila} & 256 & 8K & \textbf{64.2} & \;\,65.4$^{\dagger}$ & 57.7 & 70.1 & \;\,44.0$^{\dagger}$ & \;\,\textbf{69.7}$^{\dagger}$ & 82.2 \\
        InternVL2.5-8B~\citep{chen2024expanding} & 64 & 16K & \textbf{64.2} & - & \underline{60.0} & \;\,69.0$^{\dagger}$ & \;\,43.2$^{\dagger}$ & \;\,68.3$^{\dagger}$ & \;\,\textbf{85.0}$^{\dagger}$ \\
        Qwen2-VL-7B~\citep{wang2024qwen2} & 2fps & 16K & 63.3 & 62.3 & \;\,55.6$^{\dagger}$ & \;\,69.8$^{\dagger}$ & \;\,\underline{44.7}$^{\dagger}$ & \;\,67.9$^{\dagger}$ & \;\,81.2$^{\dagger}$ \\
        \rowcolor{lightblue} 
        SF-LLaVA-1.5-7B & 128 & 9K & \underline{63.9} & \underline{69.6} & \textbf{62.5} & \textbf{71.5} & \textbf{45.3} & \underline{68.8} & \underline{83.3} \\
        \bottomrule
    \end{tabular}
    }
    \vspace{-10pt}
    \caption{
        \textbf{Comparison with state-of-the-art models on video understanding.}
        \textbf{Bold} and \underline{underlined} are the best and second-best results for each task.
        $^{\dagger}$denotes reproduced results.
    }
    \label{tab:video_results}
    \vspace{-7pt}
\end{table}
\endgroup

\textbf{First, SF-LLaVA-1.5 achieves state-of-the-art results in long-form video understanding.}
Specifically, SF-LLaVA-1.5 outperforms existing models
on both LongVideoBench and LVBench across all model sizes.
For reference, it surpasses InternVL2.5 at both 1B (+2.3\% on LongVideoBench and +1.8\% on LVBench)
and 7B (+2.5\% on LongVideoBench and +2.1\% on LVBench) scales.
SF-LLaVA-1.5 also exhibits leading performance on MLVU.
Compared to the state-of-the-art model, Apollo, it achieves +1.0\% at the 1B scale and comparable results at other scales.
Additionally, SF-LLaVA-1.5 delivers better results even compared to Video LLMs tailored for long videos,
such as LongVU.
For instance, SF-LLaVA-1.5-3B significantly surpasses LongVU-3.2B
by +9.3\% on Video-MME and +12.9\% on MLVU.

\textbf{Second, SF-LLaVA-1.5 is the state-of-the-art model at the smaller scales.}
As edge deployment becomes increasingly important, more models are emerging in the 1B and 3B sizes,
including LLaVA-OV, InternVL2.5, Qwen2-VL, and Apollo.
For reference, SF-LLaVA-1.5-1B surpasses Qwen2-VL-2B across benchmarks
(\textit{e.g.,} 56.6\% vs. 55.6\% on Video-MME, 61.9\% vs. 53.9\% on Perception Test, 64.3\% vs. 62.7\% on MLVU).
Compared to Apollo-1.5B, SF-LLaVA-1.5-1B exhibits a +3.6\% improvement on Video-MME, while leading in other tasks.
Similarly, at the 3B scale, SF-LLaVA-1.5-3B outperforms Apollo-3B by +2.4\% on Video-MME for general Video QA
and by +1.5\% on TempCompass for temporal reasoning.

\textbf{Third, SF-LLaVA-1.5 optimizes the trade-off between performance and efficiency.}
SF-LLaVA-1.5 excels in long-form video understanding while using fewer tokens than existing methods.
Using Oryx1.5 as an example, SF-LLaVA-1.5 utilizes only $\sim$65\% of its input tokens (9K vs.\,14K)
but processes twice as many frames (128 vs.\,64), resulting in better performance on nearly all benchmarks
(\textit{e.g.,} 63.9\% vs. 58.8\% on Video-MME and 71.5\% vs. 67.5\% on MLVU).
Notably, NVILA uses a similar number of input tokens as SF-LLaVA-1.5,
yet SF-LLaVA-1.5 surpasses it by +4.8\% on LongVideoBench and +1.4\% on MLVU.
These results demonstrate the advantages of SF-LLaVA-1.5 in modeling long-range context.

\textbf{Fourth, SF-LLaVA-1.5 exhibits robustness across tasks and model sizes.}
SF-LLaVA-1.5 consistently achieves strong performance across all benchmarks in Table~\ref{tab:video_results}.
This demonstrates two key points: \textit{i)} using two-stream SlowFast inputs is beneficial for modeling long-range temporal context across various video tasks, and \textit{ii)} our proposed training pipeline and data mixture enable seamless generalization from mobile-friendly to large-scale Video LLMs.

\vspace{-5pt}
\subsection{Image Understanding Results}
\label{sec:image_results}
\vspace{-2pt}

We also compare SF-LLaVA-1.5 against recent multimodal LLMs on image understanding, as shown in Table~\ref{tab:image_results},
highlighting the following observations.

\begingroup
\setlength{\tabcolsep}{4.5pt}
\begin{table}[t]
    \centering
    \resizebox{\linewidth}{!}{%
        \begin{tabular}{l|c|c|cccc|cc|ccc}
            \toprule
            \multirow{4}{*}{Model}& \multirow{4}{*}{\makecell{Max\\Input\\Pixels}}& \multirow{4}{*}{\makecell{Train \\ Stage\\$\#$}} &\multicolumn{4}{c|}{Knowledge} &
            \multicolumn{2}{c|}{General VQA}& 
            \multicolumn{3}{c}{TextRich}\\
            \cmidrule{4-12}
            & & &\makecell{AI2D\\(test)}&\makecell{SQA\\(test)}&\makecell{MMMU\\(val)}&\makecell{MathV\\(testmini)}&MM-Vet& RW-QA & \makecell{OCRBench\\(test)} & \makecell{TextVQA\\(val)} & \makecell{DocVQA\\(test)}  \\
            \midrule
            \multicolumn{12}{c}{\textit{1B Model Comparison}}\\
            \midrule
            Gemini Nano-1~\citep{team2023gemini} &  - & - & 37.9 & - & 26.3 & 27.3 & -& -& - & 62.5& 72.2 \\
            LLaVA-OV-0.5B~\citep{li2024llava} & 5.31M & 4& 57.1& 67.2& 31.4 & 34.8& 29.1& 55.6& - &- & 70.0 \\
            MM1.5-1B~\citep{zhang2024mm1} & 4.52M & 3& 59.3&	82.1& 35.8 &	37.2& 37.4& 53.3& 60.5& \underline{72.5}& 81.0\\
            InternVL2.5-1B~\cite{chen2024expanding} &9.63M & 2& 69.3 & - & \textbf{40.9} & 43.2 & \underline{48.8}& \underline{57.5} & \textbf{78.5} & 72.0 & 84.8 \\
            MolmoE-1B~\citep{deitke2024molmo} & 4.10M& 2 &\textbf{86.4} & - & 34.9 & 34.0 & - & \textbf{60.4} & - & \textbf{78.8} & 77.7\\
            \rowcolor{isabelline} 
            SF-LLaVA-1.5-Image-1B & 2.36M & 1 & 70.8 &\textbf{87.8} & 39.3 & \textbf{51.2} & 41.1 & 57.1 & 69.5 & 70.2 & \underline{85.2} \\
            \rowcolor{lightblue} 
            SF-LLaVA-1.5-1B & 2.36M & 2 & \underline{72.8}&	\underline{87.7} & \underline{40.5} & \underline{51.0} & \textbf{51.2} & 59.2 & \underline{70.0} & 71.3 & \textbf{85.4}\\
            \midrule
            \multicolumn{12}{c}{\textit{3B Model Comparison}}\\
            \midrule
            Gemini Nano-2~\citep{team2023gemini} & - & -& 51.0 & -& 32.6& 30.6& - & -& -& 65.9& 74.3\\
            MiniCPM-V2-3B~\citep{yao2024minicpmvgpt4vlevelmllm} & 1.81M & 6 & 62.9& 80.7 & 38.2 &38.7 & 38.2 & 55.8& 60.5& \underline{74.1} & 71.9\\
            BLIP3-4B~\citep{xue2024xgen} & -& 5& -& 88.3 & 41.1 & 39.6& – & 60.5& - & 71.0& -\\
            MM1.5-3B~\citep{zhang2024mm1} & 4.52M & 3& 65.7& 85.8& 37.1& 44.4& 41.0& 56.9& 65.7& \textbf{76.5}& \underline{87.7}\\
            Phi-3.5-V-4B~\citep{abdin2024phi} & -& 3& \textbf{78.1} & \textbf{91.3} & 43.0 & 43.9 & - & -& -& 72.0 & -\\
            \rowcolor{isabelline}
            SF-LLaVA-1.5-Image-3B & 2.36M & 1 & 75.8 & 90.0	&\underline{43.7}	&\underline{57.0}
            &\textbf{51.1}	& \underline{61.8}	& \underline{72.3} & 72.0 & 87.5 \\
            \rowcolor{lightblue}
            SF-LLaVA-1.5-3B & 2.36M & 2 & \underline{77.0}& \underline{90.3} & \textbf{44.7} & \textbf{58.6} & \underline{47.5} & \textbf{63.4} & \textbf{73.4} & 73.0 & \textbf{88.8} \\
            \midrule
            \multicolumn{12}{c}{\textit{7B Model Comparison}}\\
            \midrule
            VILA1.5-8B~\citep{lin2023vila} & -& - &76.6 &- & 38.6 & 36.7 & - & 52.7&-&68.5 & 40.6 \\
            Idefics2-8B~\citep{laurencon2024matters} & 2.95M& 3 &- & -& 43.0 &51.4& - & -&-&73.0 &74.0\\
            Cambrian-1-8B~\citep{tong2025cambrian} & -& 2 &73.0 & 80.4 &42.7 &49.0 & - & 64.2 & 62.4 & 71.7 & 77.8\\
            LLaVA-OV-7B~\citep{li2024llava} & 5.31M& 4 & 81.4 & \textbf{96.0} & 48.8 & \underline{63.2} & \underline{57.5} & 66.3 & - & - & 87.5\\
            MM1.5-7B~\citep{zhang2024mm1} & 4.52M & 3 &72.2& 89.6& 41.8& 47.6& 42.2& 62.5& 63.5& 76.5& 88.1\\
            Oryx1.5-7B~\citep{liu2024oryx} & 2.36M & 3 & 79.7 & - & 47.1 & - & - &  - & 71.3 & 75.7 & 90.1 \\
            InternVL2.5-8B~\citep{chen2024expanding} & 9.63M& 2 &\textbf{84.5} &- & \textbf{56.0} & \textbf{64.4} & - & \textbf{70.1} & - & \underline{79.1} & \underline{93.0} \\
            Qwen2-VL-7B~\citep{wang2024qwen2} & -& 3&\underline{83.0} & - & \underline{54.1} & 58.2 & \textbf{62.0} & \textbf{70.1} & - & \textbf{84.3} & \textbf{94.5} \\
            \rowcolor{isabelline} 
            SF-LLaVA-1.5-Image-7B &2.36M & 1 &79.2	&\underline{91.8}	&47.0	&61.0
            &50.1	&64.6	&\underline{74.2} &75.4	& 89.7  \\
            \rowcolor{lightblue} 
            SF-LLaVA-1.5-7B &2.36M & 2 & 80.4 & 91.1 & 49.0 & 62.5 & 54.7 & \underline{67.5} & \textbf{76.4} & 76.4 & 90.3 \\
            \bottomrule
        \end{tabular}
        }
        \vspace{-7pt}
        \caption{
            \textbf{Comparison with state-of-the-art models on image understanding.}
            This table denotes ``MathV'' for MathVista and ``RW-QA'' for RealWorldQA.
            \textbf{Bold} and \underline{underlined} are the best and second-best results for each task.
        }
        \label{tab:image_results}
        \vspace{-5pt}
    \end{table}
\endgroup

\textbf{First, SF-LLaVA-1.5 excels at smaller model scales.}
Similar to video, SF-LLaVA-1.5's 1B and 3B models achieve competitive results across image benchmarks.
Specifically, SF-LLaVA-1.5-1B outperforms InternVL2.5-1B by +3.5\% on AI2D and +7.8\% on MathVista,
even though we use less than 30\% of their input resolution.
When compared to MolmoE-1B, our model clearly wins on MMMU (+5.6\%), MathVista (+17.0\%) and DocVQA (+7.7\%), although
MolmoE-1B is a specialist model optimized for image understanding.
At the 3B scale, SF-LLaVA-1.5-3B also demonstrates superior results,
(\textit{e.g.,} outperforming Phi-3.5-Vision-4B by +1.7\% on MMMU, +14.7\% on MathVista and +1.0\% on TextVQA).

\textbf{Second, SF-LLaVA-1.5 outperforms strong baselines at the 7B scale, except for InternVL2.5 and Qwen2-VL.}
Using MM1.5-7B as an example, SF-LLaVA-1.5 achieves better results across benchmarks
(\textit{e.g.,} +7.2\% on MMMU, +12.5\% on MM-Vet, and +12.9\% on OCRBench).
We are impressed by the superior results of InternVL2.5 and Qwen2-VL, especially on TextRich.
We hypothesize it is due to our \textit{(i)} lower input resolution (\textit{e.g.,} 2.36M vs.\,9.63M of InternVL2.5), \textit{(ii)} fewer training stages (\textit{e.g.,} 2 vs.\,3 of Qwen2-VL) and \textit{(iii)} frozen vision encoder.
This aligns with prior findings~\citep{zhang2024mm1} that, when the model size gets larger, higher input resolution and more training stages with fully tunable parameters are pivotal for improving the image performance.
Given that our model is video-centric and these enhancements significantly increase training costs, we leave their exploration for future work.

\textbf{Third, SF-LLaVA-1.5's image capability benefits from joint video-image training.}
SF-LLaVA-1.5, jointly optimized on video and image data, outperforms SF-LLaVA-1.5-Image on most benchmarks.
To confirm the improvements are not solely due to longer training, we conduct a second-stage training for SF-LLaVA-1.5-Image using only image data. However, the performance gap remains, indicating that joint training is the primary factor.
Additionally, the improvements are more significant on Knowledge and General benchmarks (\textit{e.g.,} +1.2\% on MMMU and +10.1\% on MM-Vet at the 1B scale).
We hypothesize this is because our video data mainly comes from lifestyle scenarios,
which could not directly benefit text-rich tasks.
A deeper analysis of joint training will be provided in Sec.~\ref{sec:video_image_ratio}.

\vspace{-5pt}
\subsection{Ablation Studies}
\label{sec:ablation_studies}
\vspace{-2pt}

All ablation studies are conducted on the 1B model with our default settings (Sec.~\ref{sec:implementation_details}).
To save training costs, models are trained on 1.2M image and 600K video samples,
randomly selected from our original data mixture (Appendix~\ref{appendix:data_mixture_details}).
The performance is evaluated on Video-MME and LongVideoBench to cover both short and long videos.

\begingroup
\setlength{\tabcolsep}{4.5pt}
\begin{table}[t]
    \scriptsize
    \centering
        \begin{tabular}{l|cccc|c}
            \toprule
            \multirow{2}{*}{Structure} & \multicolumn{4}{c|}{Video-MME (w/o sub)} & LongVideoBench \\
            & (short) & (med) & (long) & (avg) & (val) \\
            \midrule
            \rowcolor{lightblue}
            Group-based SlowFast (GSF)\quad\quad & 64.4 & \textbf{52.8} & \textbf{46.1} & \textbf{54.4} & \textbf{52.7} \\
            Interleaved SlowFast (ISF)\quad\quad & \textbf{64.7} & 52.4 & 45.3 & 54.1 & 52.3 \\
            \bottomrule
        \end{tabular}
    \vspace{-7pt}
    \caption{\textbf{Comparison between GSF and ISF on video understanding.}}
    \vspace{-5pt}
    \label{tab:gsf_vs_isf}
\end{table}
\endgroup

\vspace{-3pt}
\subsubsection{Design Choices of SlowFast}
\label{sec:slowfast_design_choices}
\vspace{-3pt}

\noindent \textbf{Group-based SlowFast (GSF) vs. Interleaved SlowFast (ISF).}
We introduced these SlowFast structures in Sec.~\ref{sec:slowfast_instantiations}
and report their video understanding results in Table~\ref{tab:gsf_vs_isf}.
GSF and ISF perform comparably on Video-MME (54.4\% vs. 54.1\% on average)
and LongVideoBench (52.7\% vs. 52.3\%),
suggesting that SF-LLaVA-1.5 is not sensitive to this design choice.
This highlights the general effectiveness of the SlowFast approach in improving long-form video understanding.
Since GSF consistently achieves superior performance across benchmarks,
we adopt it as the default SlowFast structure in this paper.

\begingroup
\setlength{\tabcolsep}{4.5pt}
\begin{table}[t]
    \scriptsize
    \centering
        \begin{tabular}{ccc|c|cccc|c}
            \toprule
            \multirow{2}{*}{\makecell{Slow Frames\\$N^{slow}$}} & \multirow{2}{*}{\makecell{Fast Frames\\$N^{fast}$}} & \multirow{2}{*}{\makecell{Total Frames\\$N$}} & \multirow{2}{*}{\makecell{Input\\Token \#}} & \multicolumn{4}{c|}{Video-MME (w/o sub)} & LongVideoBench \\
            & & & & (short) & (med) & (long) & (avg) & (val) \\
            \midrule
            32  & 0   & 32  & 7K  & 62.0 & 50.4 & 44.1 & 52.1 & 52.4 \\
            48  & 0   & 48  & 10K & \textbf{64.9} & 51.1 & 45.0 & 53.7 & 52.5 \\
            64  & 0   & 64  & 14K & 64.3 & 51.0 & 45.5 & 53.6 & 52.2 \\
            128 & 0   & 128 & 28K & 63.0 & \textbf{53.3} & 46.0 & 54.1 & 52.3 \\
            0   & 128 & 128 & 2K  & 59.3 & 49.7 & 44.3 & 51.1 & 49.7 \\
            \rowcolor{lightblue}
            32  & 128 & 128 & 9K  & 64.4 & 52.8 & \textbf{46.1} & \textbf{54.4} & \textbf{52.7} \\
            \bottomrule
        \end{tabular}
    \vspace{-7pt}
    \caption{\textbf{Results of SF-LLaVA-1.5 with different design choices on video understanding.}}
    \label{tab:slowfast_design_choices}
    \vspace{-5pt}
\end{table}
\endgroup

\noindent \textbf{Effect of the Slow and Fast Pathways.}
\textit{First}, we assess the necessity of the Slow and Fast pathways by removing them individually.
Table~\ref{tab:slowfast_design_choices} shows that SF-LLaVA-1.5 outperforms both Slow-only (row 1 vs. row 6) and Fast-only (row 5 vs. row 6) models.
This is expected since they use fewer input frames or tokens than the full model.
\textit{Second}, we test if SlowFast remains more effective
when the Slow-only model uses a comparable number of input tokens (\textit{e.g.,} 48 frames with $\sim$10K tokens).
The results (row 2 vs. row 6) demonstrate that SlowFast outperforms this baseline (\textit{e.g.,} +1.1\% on Video-MME long),
indicating that the improvements are not merely due to using more information.
\textit{Third}, we argue that SlowFast enhances both computational efficiency and long-range temporal modeling.
We verify this by comparing SlowFast with the Slow-only model that uses the same number of input frames
(\textit{e.g.,} $N^{slow}=N=128$).
The results (row 4 vs. row 6) show that SlowFast maintains superior performance while using only $\sim$30\% (9K vs. 28K) of its input tokens.

\begingroup
\setlength{\tabcolsep}{4.5pt}
\begin{table}[t]
    \scriptsize
    \centering
        \begin{tabular}{l|cc|cccc|c}
            \toprule
            \multirow{2}{*}{Video Projector} & \multirow{2}{*}{\makecell{Input\\Token \#}} & \multirow{2}{*}{\makecell{Runtime\\(per video)}} & \multicolumn{4}{c|}{Video-MME (w/o sub)} & LongVideoBench \\
            & & & (short) & (med) & (long) & (avg) & (val) \\
            \midrule
            Spatial Pooling~\citep{xu2024pllava}            & 28K & 2.40s & 63.3 & 51.8 & 45.7 & 53.6 & 51.7 \\
            Dynamic Compressor~\citep{liu2024oryx}          & 28K & 2.45s & 63.5 & 52.4 & 45.8 & 53.9 & 52.3 \\
            Qformer~\citep{li2023blip2}                     & 2K  & 1.59s & 46.7 & 43.0 & 38.4 & 42.7 & 45.0 \\
            Perceiver Resampler~\citep{jaegle2021perceiver} & 2K  & 1.50s & 52.8 & 45.9 & 43.0 & 47.2 & 48.4 \\
            \rowcolor{lightblue}
            SlowFast                                        & 9K  & 1.79s & \textbf{64.4} & \textbf{52.8} & \textbf{46.1} & \textbf{54.4} & \textbf{52.7} \\
            \bottomrule
        \end{tabular}
    \vspace{-7pt}
    \caption{
        \textbf{Comparison between SlowFast and existing video projectors on video understanding.}
        All models take 128 frames as inputs. The runtime (per video) measures only the model's forward pass
        on a single H100-80G GPU, using the LongVideoBench dataset.
    }
    \label{tab:other_video_projectors}
    \vspace{-2pt}
\end{table}
\endgroup

\noindent \textbf{SlowFast vs. Other Video Projectors.}
We compare SlowFast with existing video projectors in Table~\ref{tab:other_video_projectors}.
Specifically, we apply $2\times2$ average pooling in Spatial Pooling and Dynamic Compressor
and follow Apollo~\citep{zohar2024apollo} by using 16 tokens per frame in Q-Former and Perceiver Resampler.
All models process up to 128 input frames.
Compared to Spatial Pooling and Dynamic Compressor,
SlowFast improves runtime by 25\% while surpassing them across all benchmarks.
It also significantly outperforms Q-Former and Perceiver Resampler,
which use fixed-length tokens for information compression,
limiting their ability to handle long video sequences.
Moreover, Q-Former and Perceiver Resampler introduce additional parameters
(\textit{e.g.,} BERT-Base in Q-Former),
which restrict their advantage in runtime efficiency,
These results demonstrate SlowFast’s effectiveness
in balancing strong video performance and computational efficiency.

\subsubsection{Design Choices of Model Training}
\label{sec:video_image_ratio}

\noindent \textbf{Effect of Video-to-Image Ratio in Joint Training.}
We examine the optimal video-to-image ratio
by fixing video samples at 600K and evaluating the impact of varying image samples.
Specifically, we explore the following ratios \{1:0, 1:0.5, 1:1, 1:2, 1:3\},
where a ratio of ``1:0'' uses only video data.
Results are shown in Table~\ref{tab:video_image_ratio} with the following findings.
\textit{First}, training with only video data clearly decreases the performance in image understanding (row 1 vs. row 2),
with a substantial drop on text-rich benchmarks (\textit{e.g.,} -5.0\% on TextVQA).
\textit{Second}, joint video-image training generally improves SF-LLaVA-1.5's video capability (row 1 vs. row 4), such as on Video-MME (53.2\% vs. 54.4\% on average).
\textit{Third}, increasing the proportion of image data does not always lead to better video results (row 4 vs. row 5).
\textit{Fourth}, a video-to-image ratio of ``1:2'' achieves the best overall performance in video and image understanding,
which we adopt in our final data mixture.

\begingroup
\setlength{\tabcolsep}{4.5pt}
\begin{table}[t]
    \scriptsize
    \centering
        \begin{tabular}{l|cccc|c|cccc}
            \toprule
            \multirow{4}{*}{\;Ratio\quad\;} & \multicolumn{5}{c|}{Video Benchmarks} & \multicolumn{4}{c}{Image Benchmarks}\\
            \cmidrule{2-10}
            & \multicolumn{4}{c|}{Video-MME (w/o sub)} & LongVideoBench & MMMU & \multirow{2}{*}{RW-QA} & OCRBench & TextVQA \\
            & (short) & (med) & (long) & (avg) & (val) & (val) & & (test) & (val) \\
            \midrule
            \;1 : 0 & 63.4 & 51.8 & 44.3 & 53.2 & 52.0 & 39.4 & 55.8 & 61.6 & 64.2 \\
            \;1 : 0.5 & 65.1 & 50.1 & 45.9 & 53.7 & 52.3 & 44.0 & 59.0 & 66.2 & 69.2 \\
            \;1 : 1 & \textbf{64.8} & 50.5 & 45.3 & 53.5 & 52.1 & 39.9 & 58.5 & \textbf{68.3}  & 69.5 \\
            \rowcolor{lightblue}
            \;1 : 2 & 64.4 & \textbf{52.8} & \textbf{46.1} & \textbf{54.4} & \textbf{52.7} & 40.0 & \textbf{59.1} & 68.2 & \textbf{69.7} \\
            \;1 : 3 & 63.7 & 52.3 & 46.0 & 54.0 & 52.5 & \textbf{40.7} & 58.8 & 68.0 & 69.3 \\
            \bottomrule
        \end{tabular}
    \vspace{-7pt}
    \caption{
        \textbf{Results of using different video-to-image data ratios in joint training.}
    }
    \label{tab:video_image_ratio}
    \vspace{-2pt}
\end{table}
\endgroup

\vspace{-7pt}
\section{Limitations}
\label{sec:limitations}
\vspace{-7pt}

\textit{First}, SF-LLaVA-1.5 prefers FPS sampling,
but falls back to uniform sampling
when the video duration exceeds the maximum frame capacity (\textit{i.e.,} 128 in this paper).
This approach may miss some key frames in long-form videos and mislead the model about a video's playback speed
(\textit{e.g.,} A ten-minute video and a one-hour video have the same number of input frames).
Developing an efficient memory model to summarize the long-range context
is a promising direction~\citep{xu2021long}.
We can also input extra information (\textit{e.g.,} frame timestamps) to enhance the temporal modeling.
\textit{Second},
SF-LLaVA-1.5's performance can be further improved by tuning all parameters, including the visual encoder.
However, we found this is not trivial for Long Video LLMs due to the high GPU memory cost of caching the activation values.
Future studies could explore the integration of memory-saving techniques, such as Stochastic BP~\citep{cheng2022stochastic}.
More analysis will be discussed in Appendix~\ref{appendix:more_ablation_studies}.

\vspace{-7pt}
\section{Conclusion}
\label{sec:conclusion}
\vspace{-7pt}

Building upon the insights of SlowFast-LLaVA~\citep{xu2024slowfast},
we introduce SlowFast-LLaVA-1.5, a new family of token-efficient Video LLMs for long-form video understanding.
While SlowFast-LLaVA adapts the two-stream SlowFast inputs into a training-free model,
this work explores further improvements by building a supervised fine-tuning pipeline with a high-quality data mixture.
Our model family, ranging from 1B to 7B parameters, focuses on developing lightweight models that are both compact for potential edge deployment and powerful for various video tasks.
Experimental results demonstrate that SlowFast-LLaVA-1.5 achieves superior performance
across video benchmarks while maintaining strong image capabilities.
We hope our work inspires the community to develop efficient yet robust Long Video LLMs based on open-source datasets.

\section*{Acknowledgments}
\label{sec:acknowledgments}

We thank Yizhe Zhang, Feng Tang, Jesse Allardice, Jiaming Hu, Yihao Qian, Zhe Fu, Hong-You Chen, Wentao Wu, Junting Pan, Bowen Zhang, Yanghao Li for their kind help.

\bibliography{colm2025_conference}
\bibliographystyle{colm2025_conference}

\clearpage
\appendix
\section{Appendix}
\label{sec:appendix}

\subsection{Details of Data Mixture}
\label{appendix:data_mixture_details}

\vspace{-7pt}
\begingroup
\setlength{\tabcolsep}{4.5pt}
\begin{table}[!h]
    \centering
    \resizebox{1.0\linewidth}{!}{%
    \begin{tabular}{c|c|c|c}
    \toprule
         \textbf{Mixture}& \textbf{Data Category}& \textbf{Datasets} & \textbf{\# Samples} \\
         \midrule
         \multirow{35}{*}{\textbf{Image Mixture}}& \textbf{General}& \makecell{LLaVA  Complex Reasoning~\citep{liu2023llava}, \\ LLaVA Conversation~\citep{liu2023llava}, \\ ShareGPT-4v~\citep{chen2024sharegpt4v}, Coco Caption~\citep{chen2015microsoft}, \\ LLaVA v1.5 VQAv2 OKVQA~\citep{liu2023improvedllava}, \\ LLaVA v1.5 GQA~\citep{liu2023improvedllava}, \\LLaVA v1.5 A-OKVQA~\citep{liu2023improvedllava}, \\ Pixmo-Ask-Model-Anything~\citep{deitke2024molmo}, \\ Image Textualization~\citep{pi2024image}, ShareGPT4o~\citep{cui2025comprehensive}, \\Vision FLAN~\citep{xu2024vision}, VizWiz~\citep{gurari2018vizwiz}, \\ TallyQA~\citep{acharya2019tallyqa}, Visual7W~\citep{zhu2016visual7w}, \\ VQARAD~\citep{lau2018dataset_vqarad}, VSR~\citep{liu2023visual_vsr}, \\ Hateful Memes~\citep{kiela2020hateful}} & \multirow{35}{*}{4.67M}\\
         \cmidrule{2-3}
                     & \textbf{TextRich} & \makecell{OCRVQA~\citep{mishraICDAR19_OCRVQA}, Synthdog-En~\citep{kim2022donut_synthdog}, \\ TextCaps~\citep{sidorov2019textcaps}, TextVQA~\citep{singh2019towards}, \\ DVQA~\citep{kafle2018dvqa}, ChartQA~\citep{masry-etal-2022-chartqa}, \\ DocVQA~\citep{mathew2021docvqa}, InfoVQA~\citep{mathew2022infographicvqa}, \\ VisualMRC~\citep{VisualMRC2021}, WikiTQ~\citep{pasupat2015compositional_wtq}, \\ DeepForm~\citep{svetlichnaya2020deepform}, KleisterCharity~\citep{stanislawek2021kleister},  \\ TabFact~\citep{2019TabFactA}, ScreenQA~\citep{baechler2024screenai}, \\ TabMWP~\citep{lu2022dynamic_tabmwp}, ST-VQA~\citep{biten2019scene}, \\ VisText~\citep{2023-vistext}, HiTab~\citep{cheng2021hitab}, \\ArxivQA~\citep{li2024multimodal_arxivqa}, WikiSQL~\citep{zhongSeq2SQL2017}, \\ Chart2Text~\citep{obeid2020chart}, RenderedText~\citep{render_text_hf}, \\ FinQA~\citep{chen2021finqa}, TAT-QA~\citep{zhu2021tat}, \\ Pixmo-Docs~\citep{deitke2024molmo}, PlotQA~\citep{Methani_2020_WACV_plotqa}, \\ MMC-Instruct~\citep{liu2023mmc}, ArT~\citep{zhang2022arbitrary}, \\ NAF~\citep{davis2019deep_naf}, SROIE~\citep{huang2019icdar2019_sroie}, \\ LRV Chart~\citep{liu2023aligning_lrv}, FigureQA~\citep{kahou2017figureqa}, \\ RoBUT SQA~\citep{han2023robustqa}, Screen2Words~\citep{wang2021screen2words}, \\ HME100K~\citep{yuan2022syntax_hme100k}, UReader~\citep{ye2023ureader}, \\ Diagram Image2Text~\citep{laurencon2024matters}, ChromeWriting~\citep{mouchere2011crohme2011}, \\ IIIT5K~\citep{MishraBMVC12_iiit5k}, IAM~\citep{marti2002iam}, \\ TextOCR~\citep{singh2021textocr}, K12 Printing~\citep{k12_hf}} & \\
                     \cmidrule{2-3}
                     &  \textbf{Kowledge}&\makecell{AI2D~\citep{kembhavi2016diagram}, ScienceQA~\citep{lu2022learn}, \\GeomVerse~\citep{kazemi2023geomverse}, CLEVER~\citep{johnson2017clevr}, \\IconQA~\citep{lu2021iconqa}, RAVEN~\citep{zhang2019raven}, \\ Inter-GPS~\citep{lu2021inter}, WebSight~\citep{laurencon2024unlocking_websight},\\ DaTikZ~\citep{belouadi2024detikzify}, Design2Code~\citep{si2024design2code}, \\ TQA~\citep{kembhavi2017you_tqa}, MAVIS MCollect~\citep{zhang2024mavismathematicalvisualinstruction,li2024llava}, \\ MAVIS Data Engine~\citep{zhang2024mavismathematicalvisualinstruction,li2024llava}, Geo170K~\citep{gao2023g_geo170k}, \\ Geo170K Align~\citep{gao2023g_geo170k,li2024llava}, Geometry3K~\citep{lu2021inter}, \\ GEOS~\citep{seo2015solving_geos}, GeoQA+~\citep{cao2022augmented_geoqaplus}, \\ MapQA~\citep{chang2022mapqa}, Super-CLEVR~\citep{li2023super}, \\ UniGeo~\citep{chen2022unigeo}} \\
        \midrule
        \multirow{1}{*}{\textbf{Video Mixture}} & \textbf{General} & \makecell{LLaVA-Hound~\citep{zhang2024direct}, ShareGPT4Video~\citep{chen2024sharegpt4video}, \\ VideoChatGPT-Plus~\citep{Maaz2024VideoGPT+}, LLaVA-Video-178K~\citep{llava178k}, \\ Cinepile~\citep{rawal2024cinepile}, ActivityNet-QA~\citep{yu2019activityqa}, \\ NExT-QA~\citep{xiao2021next}, Perception Test~\citep{patraucean2023perception}} & \multirow{1}{*}{2.01M} \\
        \bottomrule
    \end{tabular}
    }
    \vspace{-10pt}
    \caption{\textbf{Details of our image and video mixtures.}}
    \label{tab:data_mixture}
\end{table}
\endgroup

\vspace{-10pt}
\subsection{Benchmarks and Metrics}
\label{appendix:benchmarks_and_metrics}

All evaluations are performed using the \texttt{lmms-eval}\footnote{https://github.com/EvolvingLMMs-Lab/lmms-eval.} toolkit, where we use the official evaluation metrics to report numbers without any filtering on the prediction outputs.

\begingroup
\setlength{\tabcolsep}{4.5pt}
\begin{table}[h]
    \scriptsize
    \centering
        \begin{tabular}{ll|ccc}
            \toprule
            Category & Benchmark\quad\quad & \# Videos & \# QAs & Avg Duration (s) \\
            \midrule
            \multirow{4}{*}{General Video QA} & Video-MME~\citep{fu2024video} & 900 & 2700 & 1010 \\
                                              & Perception Test (val)~\citep{patraucean2023perception} & 5900 & 19139 & 23 \\
                                              & ActivitiyNet-QA (test)~\citep{yu2019activityqa} & 800 & 8000 & 180 \\
                                              & VCGBench (test)~\cite{Maaz2023VideoChatGPT} & 800 & 3497 & 180 \\
            \midrule
            \multirow{3}{*}{Long-Form Video Understanding} & LongVideoBench (val)~\citep{wu2025longvideobench} & 752 & 1337 & 473 \\
                                                           & MLVU (test)~\citep{zhou2024mlvu} & 1730 & 3102 & 930 \\
                                                           & LVBench (test)~\citep{wang2024lvbench} & 103 & 1549 & 4101 \\
            \midrule
            \multirow{2}{*}{Temporal Reasoning} & TempCompass (mc)~\citep{liu2024tempcompass} & 410 & 7540 & - \\
                                                & NExT-QA (mc)~\citep{xiao2021next} & 1000 & 8564 & 44 \\
            \bottomrule
        \end{tabular}
    \vspace{-7pt}
    \caption{
        \textbf{Details of video understanding benchmarks}.
    }
    \label{tab:video_benchmarks}
\end{table}
\endgroup

\subsubsection{Video Benchmarks}
\label{appendix:video_benchmarks}

We evaluate our model on various video understanding benchmarks in Table~\ref{tab:video_benchmarks}.

\subsubsection{Image Benchmarks}
\label{appendix:image_benchmarks}

We evaluate our model on the following image understanding benchmarks:
\begin{itemize}[leftmargin=22pt]
    \item \textbf{Knowledge Image QA} inspects a model's capability of answering questions requiring knowledge in specific domains. Our model is evaluated on AI2D~\citep{kembhavi2016diagram} and ScienceQA~\citep{lu2022learn} for science, MathVISTA~\cite{lu2024mathvista} for math and MMMU~\citep{yue2023mmmu} for multi-discipline tasks.
    \item \textbf{General Image QA} evaluates the general image capability of our model. We select RealWorldQA\footnote{https://huggingface.co/datasets/xai-org/RealworldQA} and MMVet~\citep{yu2024mmvet} to serve this purpose, where RealWorldQA examines a model's capability in real-world scenarios and MMVet assesses a model's performance for more complicated tasks.
    \item \textbf{TextRich Image QA} contains images embeded with dense texts. To achieve high performance, a model is expected to excel at reasoning over reading. We include OCRBench~\citep{Liu_2024_ocrbench}, TextVQA~\citep{singh2019towards} and DocVQA~\citep{mathew2021docvqa} measuring OCR, scene text and document understanding, respectively.
\end{itemize}

\subsubsection{Instantiations of SlowFast Cont’d}
\label{appendix:slowfast_instantiations_cont}
\vspace{-5pt}

\begin{figure*}[h]
    \centering
    \includegraphics[width=1.0\textwidth]{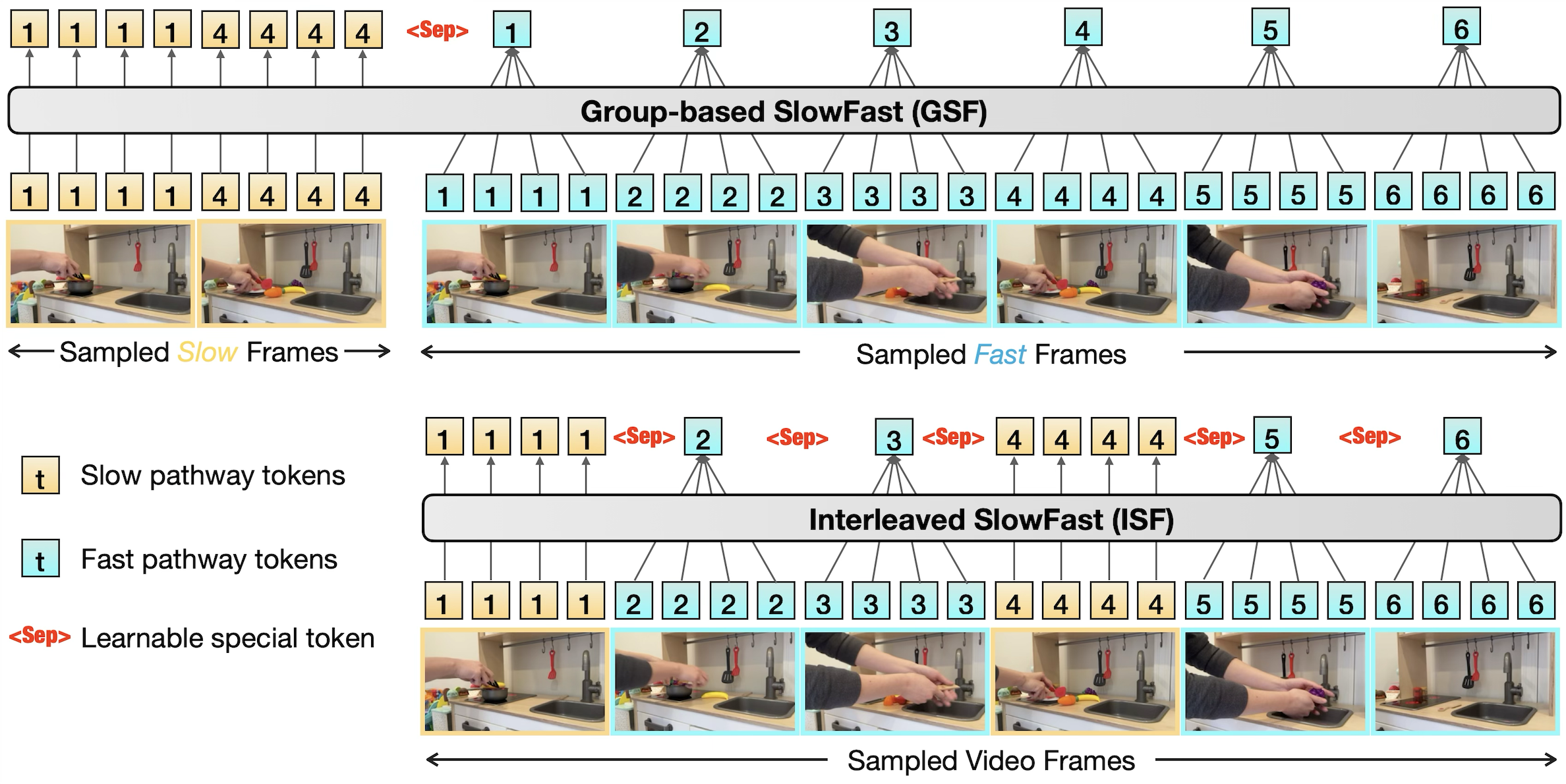}
    \vspace{-22pt}
    \caption{
        \textbf{Visualization of Group-based SlowFast (GSF) and Interleaved SlowFast (ISF).}
        In this example, each sliding window contains three frames, with the first frame serving as Slow (color yellow) and the others as Fast (color cyan). The number of each token indicates the timestamp of the frame it corresponds to. (Best viewed in color.)
    }
    \label{fig:gsf_vs_isf}
\end{figure*}

\vspace{-5pt}
\subsection{More Video Understanding Results}
\label{appendix:more_video_results}
\vspace{-2pt}

We compare with recent Video LLMs as representative examples in Table~\ref{tab:video_results},
and here, we include a broader group of models in Table~\ref{tab:more_video_results}.
For ActivityNet-QA and VCGBench, we adopt the GPT-assisted evaluation to assess the accuracy.
Specifically, we use \texttt{GPT-3.5-Turbo-0125} as the judge.
\textbf{It is worth noting that} our model cannot be directly compared with previous work
that uses \texttt{GPT-3.5-Turbo-0613} (deprecated by OpenAI) or an unknown version,
since different GPT versions can significantly impact the results~\citep{FreeVA}.

\begin{table}[t]
    \scriptsize
    \centering
    \resizebox{\linewidth}{!}{%
    \begin{tabular}{l|ccccc|ccc|cc}
        \toprule
        \multirow{10}{*}{Model} & \multicolumn{5}{c|}{General VideoQA} & \multicolumn{3}{c|}{\makecell{Long-Form Video\\Understanding}} & \multicolumn{2}{c}{\makecell{Temporal\\Reasoning}} \\
        \cmidrule{2-11}
        & \rotatebox{90}{VideoMME} & \rotatebox{90}{VideoMME} & \rotatebox{90}{PercepTest} & \rotatebox{90}{ActivityNet-QA} & \rotatebox{90}{VCGBench} & \rotatebox{90}{LongVideoBench} & \rotatebox{90}{MLVU} & \rotatebox{90}{LVBench} & \rotatebox{90}{TempComp} & \rotatebox{90}{NExT-QA} \\
        & (w/o sub) & (w/ sub) & (val) & (test) & (test) & (val) & (dev) & (avg) & (mc) & (test) \\
        \midrule
        \multicolumn{11}{c}{\textit{Proprietary Models}}\\
        \midrule
        GPT-4V~\citep{gpt4v} & 59.9 & 63.3 & - & 57.0 & 4.06 & 61.3 & 49.2 & - & -  & - \\
        GPT-4o~\citep{gpt4o} & 71.9 & 77.2 & - & - & - & 66.7 & 64.6 & 30.8 & 70.9 & - \\
        Gemini-1.5-Flash~\citep{team2023gemini} & 70.3 & 75.0 & - & - & - & 61.6 & - & - & - & - \\
        Gemini-1.5-Pro~\citep{team2023gemini} & 75.0 & 81.3 & - & 57.5 & - & 64.0 & - & 33.1 & 69.3 & - \\
        \midrule
        \multicolumn{11}{c}{\textit{1B Model Comparison}}\\
        \midrule
        LLaVA-OV-0.5B~\citep{li2024llava} & 44.0 & 43.5 & 49.2 & \;\,\textcolor{lavendergray}{50.5$^{\ddagger}$} & \;\,\textcolor{lavendergray}{3.12$^{\ddagger}$} & 45.8 & 50.3 & \;\,32.7$^{\dagger}$ & \;\,53.2$^{\dagger}$ & 57.2  \\
        MM1.5-1B~\citep{zhang2024mm1} & 45.7 & - & - & \textbf{56.1} & \underline{3.14} & 43.9 & - & - & - & 71.8 \\
        Apollo-1.5B~\citep{zohar2024apollo} & 53.0 & 54.6 & \underline{61.0} & - & - & \underline{54.1} & \underline{63.3} & - & \textbf{60.8} & - \\
        LinVT-Mipha-1.6B~\citep{gao2024linvt} & 44.5 & 46.1 & - & \;\,\textcolor{lavendergray}{47.5$^{\ddagger}$} & - & 49.7 & 56.2 & - & 45.2 & 71.1 \\
        InternVL2.5-2B~\citep{chen2024expanding} & 51.9 & 54.1 & - & - & - & 52.0 & 61.4 & \;\,37.9$^{\dagger}$ & \;\,53.4$^{\dagger}$ & \;\,\textbf{77.2}$^{\dagger}$ \\
        Qwen2-VL-2B~\citep{wang2024qwen2} & \underline{55.6} & \textbf{60.4} & 53.9 & - & - & \;\,48.7$^{\dagger}$ & \;\,62.7$^{\dagger}$ & \;\,\underline{39.4}$^{\dagger}$ & \;\,\underline{60.6}$^{\dagger}$ & \;\,\textbf{77.2}$^{\dagger}$ \\
        \rowcolor{lightblue} 
        SF-LLaVA-1.5-1B & \textbf{56.6} & \underline{58.1} & \textbf{61.9} & \underline{52.9} & \textbf{3.27} & \textbf{54.3} & \textbf{64.3} &  \textbf{39.7} & 60.5 & \underline{76.7} \\
        \midrule
        \multicolumn{11}{c}{\textit{3B Model Comparison}}\\
        \midrule
        Blip3-Video-4B~\citep{ryoo2024xgen} & - & - & - & \;\,\textcolor{lavendergray}{56.9$^{\ddagger}$} & - & - & - & - & - & 77.1 \\
        Phi-3.5-V-4B~\citep{abdin2024phi} & 51.5 & - & - & - & - & - & - & - & - & - \\
        V-Ma$^2$mba-3.1B~\citep{lee2024look} & 45.2 & - & - & 51.7 & 3.03 & 43.0 & - & - & - & - \\
        VILA1.5-3B~\citep{liu2024nvila} & 42.2 & 44.2 & 49.1 & \;\,\textcolor{lavendergray}{50.7$^{\ddagger}$} & - & 42.9 & 44.4 & - & 56.1 & - \\
        MM1.5-3B~\citep{zhang2024mm1} & 49.5 & - & - & \textbf{57.9} & \underline{3.17} & 45.4 & - & - & - & 74.7 \\
        LongVU-3.2B~\citep{shen2024longvu} & 51.5 & - & - & - & - & - & 55.9 & - & - & - \\
        InternVL2-4B~\citep{chen2024far} & 53.9 & 57.0 & \;\,53.9$^{\dagger}$ & - & - & 53.0 & 59.9 & \;\,\underline{35.1}$^{\dagger}$ & \;\,60.2$^{\dagger}$ & \;\,71.1$^{\dagger}$ \\
        LinVT-Blip3-4B~\citep{zohar2024apollo} & 58.3 & \underline{62.4} & - & \;\,\textcolor{lavendergray}{58.9$^{\ddagger}$} & - & \underline{56.6} & 67.9 & - & 59.6 & \underline{80.1} \\
        Apollo-3B~\citep{zohar2024apollo} & \underline{58.4} & 60.6 & \underline{65.0} & - & - & 55.1 & \underline{68.7} & - & \underline{62.5} & -\\
        \rowcolor{lightblue} 
        SF-LLaVA-1.5-3B & \textbf{60.8} & \textbf{63.1} & \textbf{65.8} & \underline{55.5} & \textbf{3.32} & \textbf{57.3} & \textbf{68.8} & \textbf{43.3} & \textbf{64.0} & \textbf{80.8} \\
        \midrule
        \multicolumn{11}{c}{\textit{7B Model Comparison}}\\
        \midrule
        VideoChatGPT-7B~\citep{Maaz2023VideoChatGPT} & - & - & - & 35.2 & 2.42 & - & - & - & \,\,43.5$^*$ & - \\
        VideoLLaVA-7B~\citep{lin2023video} & \,\,39.9$^*$ & 41.6 & - & 45.3 & - & \,\,39.1$^*$ & \,\,47.3$^*$ & - & \,\,49.8$^*$ & - \\
        MovieChat+-7B~\citep{song2024moviechat} & - & - & - & \;\,\textcolor{lavendergray}{48.1$^{\ddagger}$} & \;\,\textcolor{lavendergray}{2.73$^{\ddagger}$} & - & - & 22.5$^*$ & - & 54.8 \\
        PLLaVA-7B~\citep{xu2024pllava} & - & - & - & 56.3 & 3.12 & \,\,40.2$^*$ & - & - & - & - \\
        Tarsier-7B~\citep{wang2024tarsier} & - & - & - & \underline{59.5} & - & - & - & - & - & 71.6 \\
        LLaVA-Next-Video-7B~\citep{zhang2024llavanextvideo} & - & - & - & \;\,\textcolor{lavendergray}{53.5$^{\ddagger}$} & \;\,\textcolor{lavendergray}{3.26$^{\ddagger}$} & - & - & - & - & - \\
        VideoChat2-HD-7B~\citep{li2023mvbench} & 45.3 & 55.7 & 47.3 & - & - & 3.10 & - & - & \,\,48.8$^*$ & 79.5 \\
        VideoLLaMA2-7B~\citep{damonlpsg2024videollama2} & 47.9 & 50.3 & 51.4 & \;\,\textcolor{lavendergray}{50.2$^{\ddagger}$} & \;\,\textcolor{lavendergray}{3.13$^{\ddagger}$} & - & \,\,48.5$^*$ & - & - & - \\
        VideoCCAM-9B~\citep{fei2024video} & 53.9 & 56.1 & - & \;\,\textcolor{lavendergray}{59.7$^{\ddagger}$} & - & - & 63.1 & - & - & - \\
        Flash-VStream-7B~\citep{zhang2024flash} & - & - & - & \;\,\textcolor{lavendergray}{51.9$^{\ddagger}$} & - & - & - & - & - & 61.6 \\
        VILA-1.5-8B~\citep{lin2023vila} & - & - & 41.8 & \;\,\textcolor{lavendergray}{54.3$^{\ddagger}$} & - & - & - & - & \,\,58.8$^*$ & - \\
        TimeMaker-8B~\citep{chen2024timemarker} & 57.3 & - & - & - & - & 56.3 & 49.2 & 41.3  & 60.4 & - \\
        LongVA-7B~\citep{zhang2024long} & 52.6 & 54.3 & - & - & \;\,\textcolor{lavendergray}{3.57$^{\ddagger}$} & - & 56.3 & - & \,\,57.0$^*$ & 69.3 \\
        LongVILA-7B~\citep{chen2024longvila} & 60.1 & 65.1 & 58.1 & \;\,\textcolor{lavendergray}{59.5$^{\ddagger}$} & - & 57.1 & - & - & - & 80.7 \\
        LongVU-7B~\citep{shen2024longvu} & 60.6 & - & - & - & - & - & 65.4 & - & - & - \\
        XComposer-8B~\citep{zhang2024internlm} & 55.8 & 58.8 & 34.4 & - & - & - & 37.3 & - & \,\,62.1$^*$ & - \\
        VideoLLaMA2.1-7B~\citep{damonlpsg2024videollama2} & 54.9 & 56.4 & 54.9 & \;\,\textcolor{lavendergray}{53.0$^{\ddagger}$} & - & - & 57.4 & 36.2 & 56.8 & 75.6 \\
        LinVT-Qwen2-VL-7B~\citep{gao2024linvt} & 63.1 & 63.3 & - & \;\,\textcolor{lavendergray}{60.1$^{\ddagger}$} & - & 57.2 & 68.9 & - & 65.8 & \textbf{85.5} \\
        MM1.5-7B~\citep{zhang2024mm1} & 53.5 & - & - & \textbf{60.9} & \underline{3.22} & 49.4 & - & - & - & 76.9 \\
        Kangaroo-8B~\citep{liu2024kangaroo} & 56.0 & 57.6 & - & - & - & 54.8 & 61.0 & 39.4 & 62.5 & - \\
        Oryx1.5-7B~\citep{liu2024oryx} & 58.8 & 64.2 & \textbf{70.0} & - & \;\,\textcolor{lavendergray}{3.62$^{\ddagger}$} & 56.3 & 67.5 & \;\,39.0$^{\dagger}$ & \;\,58.8$^{\dagger}$ & 81.8 \\
        LLaVA-OV-7B~\citep{li2024llava} & 58.2 & 61.5 & 49.7 & \;\,\textcolor{lavendergray}{56.6$^{\ddagger}$} & \;\,\textcolor{lavendergray}{3.51$^{\ddagger}$} & 56.5 & 64.7 & - & \;\,64.2$^{\dagger}$ & 79.4 \\
        LLaVA-Video-7B~\citep{llava178k} & 63.3 & \underline{69.7} & 66.9 & \;\,\textcolor{lavendergray}{56.5$^{\ddagger}$} & \;\,\textcolor{lavendergray}{3.52$^{\ddagger}$} &  58.2 & 70.8 & - & - & 83.2 \\
        Apollo-7B~\citep{zohar2024apollo} & 61.3 & 63.3 & 67.3 & - & - & 58.5 & \underline{70.9} & - & 64.9 & - \\
        NVILA-8B~\citep{liu2024nvila} & \textbf{64.2} & \textbf{70.0} & \;\,65.4$^{\dagger}$ & \textbf{60.9} & - & 57.7 & 70.1 & \;\,44.0$^{\dagger}$ & \;\,\textbf{69.7}$^{\dagger}$ & 82.2 \\
        InternVL2.5-8B~\citep{chen2024expanding} & \textbf{64.2} & 66.9 & - & - & - & \underline{60.0} & \;\,69.0$^{\dagger}$ & \;\,43.2$^{\dagger}$ & \;\,68.3$^{\dagger}$ & \;\,\underline{85.0}$^{\dagger}$ \\
        Qwen2-VL-7B~\citep{wang2024qwen2} & 63.3 & 69.0 & 62.3 & - & - & \;\,55.6$^{\dagger}$ & \;\,69.8$^{\dagger}$ & \;\,\underline{44.7}$^{\dagger}$ & \;\,67.9$^{\dagger}$ & \;\,81.2$^{\dagger}$ \\
        \rowcolor{lightblue} 
        SF-LLaVA-1.5-7B & \underline{63.9} & 65.4 & \underline{69.6} & 57.0 & \textbf{3.35} & \textbf{62.5} & \textbf{71.5} & \textbf{45.3} & \underline{68.8} & 83.3 \\
        \bottomrule
    \end{tabular}
    }
    \vspace{-5pt}
    \caption{
        \textbf{Comparison with a broader group of Video LLMs on video understanding.}
        $^{\dagger}$denotes reproduced results. $^*$denotes results from the benchmark leaderboard.
        $^{\ddagger}$denotes results evaluated using \texttt{GPT-3.5-Turbo-0613} or an unknown version,
        which cannot be directly compared with our results.
        \textbf{Bold} and \underline{underlined} are the best and second-best results for each task.
    }
    \label{tab:more_video_results}
\end{table}

\vspace{-2pt}
\subsection{Effect of Training the Visual Encoder}
\label{appendix:more_ablation_studies}

By default, the visual encoder is frozen in both Stage I and II.
We now assess whether training the visual encoder improves the image and video understanding performance.

We start with training Stage I, tuning the visual encoder together with other parameters
(named as \textbf{SF-LLaVA-1.5-Image-E2E}).
We evaluate it on image benchmarks, with results presented in Table~\ref{tab:e2e_image_results}.
We observe that training the visual encoder significantly improves the image performance,
especially on Text-Rich tasks (row 1 and row 2 of each model scale).
For reference, SF-LLaVA-1.5-Image-E2E-3B outperforms SF-LLaVA-1.5-Image-3B by +4.9\% on OCRBench and +2.7\% on TextVQA.

We move on to Stage II with fully tunable parameters but encounter the out-of-memory issue
(even when we train the 1B model with batch size 1 on H100-80G GPUs).
This issue arises from caching a large number of activation values from the visual encoder while extracting features from 128 input frames --- that is why we do not have this problem in Stage I.
Stochastic BP~\citep{cheng2022stochastic} is proposed to solve this problem and is utilized by modern temporal action detectors~\citep{cheng2022tallformer} for efficient end-to-end training.
However, integrating this memory-saving technique into multimodal LLMs is non-trivial and is left for future exploration.

Finally, we test if tuning the visual encoder only in Stage I and freezing it in Stage II is effective. We train models (named as \textbf{SF-LLaVA-1.5-E2E}) based on SF-LLaVA-1.5-Image-E2E, with the visual encoder frozen.
The models are evaluated on both image and video benchmarks, as shown in Table~\ref{tab:e2e_image_results} and Table~\ref{tab:e2e_video_results}. The results show that SF-LLaVA-1.5-E2E performs significantly worse than SF-LLaVA-1.5 across all metrics. We argue that tuning the visual encoder in Stage I harms its generalization ability, leading to overfitting on image tasks and conflicts between image and video tasks. We will explore the optimal training strategy for Video LLMs in future work.

\begingroup
\setlength{\tabcolsep}{4.5pt}
\begin{table}[t]
    \centering
    \resizebox{\linewidth}{!}{%
        \begin{tabular}{l|cc|cccc|cc|ccc}
            \toprule
            \multirow{4}{*}{Model}& \multicolumn{2}{c|}{\makecell{Training\\Visual Encoder}} &\multicolumn{4}{c|}{Knowledge} & \multicolumn{2}{c|}{General VQA} & \multicolumn{3}{c}{Text-Rich}\\
            \cmidrule{2-12}
            & Stage I & Stage II &\makecell{AI2D\\(test)}&\makecell{SQA\\(test)}&\makecell{MMMU\\(val)}&\makecell{MathV\\(testmini)}&MM-Vet& RW-QA & \makecell{OCRBench\\(test)} & \makecell{TextVQA\\(val)} & \makecell{DocVQA\\(test)}  \\
            \midrule
            \multicolumn{12}{c}{\textit{1B Model Comparison}}\\
            \midrule
            SF-LLaVA-1.5-Image-E2E-1B & \ding{52} & - & \textbf{73.9} & \textbf{89.3} & 38.3 & \textbf{53.0} & 41.1 & \textbf{60.3} & \textbf{74.0} & \textbf{73.8} & \textbf{87.8} \\
            \rowcolor{isabelline}
            SF-LLaVA-1.5-Image-1B & \textcolor{red}{\ding{56}} & - & 70.8 & 87.8 & 39.3 & 51.2 & 41.1 & 57.1 & 69.5 & 70.2 & 85.2 \\
            SF-LLaVA-1.5-E2E-1B & \ding{52} & \textcolor{red}{\ding{56}} & 70.5 & 81.7 & 38.9 & 41.7 & 34.3 & 55.7 & 48.8 & 60.1 & 68.1 \\
            \rowcolor{lightblue}
            SF-LLaVA-1.5-1B & \textcolor{red}{\ding{56}} & \textcolor{red}{\ding{56}} & 72.8 & 87.7 & \textbf{40.5} & \textbf{51.0} & 51.2 & 59.2 & 70.0 & 71.3 & 85.4 \\
            \midrule
            \multicolumn{12}{c}{\textit{3B Model Comparison}}\\
            \midrule
            SF-LLaVA-1.5-Image-E2E-3B & \ding{52} & - & \textbf{77.2} & 90.0 & 44.1 & \textbf{61.1} & 48.0 & 61.8 & \textbf{77.2} & \textbf{74.7} & \textbf{90.0} \\
            \rowcolor{isabelline}
            SF-LLaVA-1.5-Image-3B & \textcolor{red}{\ding{56}} & - & 75.8 & 90.0 & 43.7 & 57.0 & \textbf{51.1} & 61.8 & 72.3 & 72.0 & 87.5 \\
            SF-LLaVA-1.5-E2E-3B & \ding{52} & \textcolor{red}{\ding{56}} & 75.2 & 84.3 & 44.2 & 47.8 & 38.6 & 56.9 & 51.6 & 64.9 & 72.9 \\
            \rowcolor{lightblue}
            SF-LLaVA-1.5-3B & \textcolor{red}{\ding{56}} & \textcolor{red}{\ding{56}} & 77.0 & \textbf{90.3} & \textbf{44.7} & 58.6 & 47.5 & \textbf{63.4} & 73.4 & 73.0 & 88.8 \\
            \midrule
            \multicolumn{12}{c}{\textit{7B Model Comparison}}\\
            \midrule
            SF-LLaVA-1.5-Image-E2E-7B & \ding{52} & - & 79.5 & 91.2 & 47.1 & \textbf{63.5} & 47.4 & 66.9 & \textbf{78.3} & 75.8 & \textbf{90.7} \\
            \rowcolor{isabelline}
            SF-LLaVA-1.5-Image-7B & \textcolor{red}{\ding{56}} & - & 79.2 & \textbf{91.8} & 47.0 & 61.0 & 50.1 & 64.6 & 74.2 & 75.4 & 89.7 \\
            SF-LLaVA-1.5-E2E-7B & \ding{52} & \textcolor{red}{\ding{56}} & 76.7 & 85.8 & 44.4 & 54.0 & 44.9 & 60.5 & 59.6 & 70.8 & 78.8 \\
            \rowcolor{lightblue}
            SF-LLaVA-1.5-7B & \textcolor{red}{\ding{56}} & \textcolor{red}{\ding{56}} & \textbf{80.4} & 91.1 & \textbf{49.0} & 62.5 & \textbf{54.7} & \textbf{67.5} & 76.4 & \textbf{76.4} & 90.3 \\
            \bottomrule
        \end{tabular}
        }
        \vspace{-5pt}
        \caption{
            \textbf{Results of SF-LLaVA-1.5-E2E and SF-LLaVA-1.5-Image-E2E on image benchmarks},
            which fully train the visual encoder together with the projector and LLM.
        }
        \label{tab:e2e_image_results}
        \vspace{15pt}
    \end{table}
\endgroup

\begingroup
\setlength{\tabcolsep}{4.5pt}
\begin{table}[t]
    \centering
    \resizebox{\linewidth}{!}{%
    \begin{tabular}{l|cc|cc|ccc|cc}
        \toprule
        \multirow{4}{*}{Model} & \multicolumn{2}{c|}{\makecell{Training\\Visual Encoder}} & \multicolumn{2}{c|}{General VideoQA} & \multicolumn{3}{c|}{Long-Form Video Understanding} & \multicolumn{2}{c}{Temporal Reasoning} \\
        \cmidrule{2-10}
        & Stage I & Stage II & \makecell{VideoMME\\(w/o sub)} & \makecell{PercepTest\\(val)} & \makecell{LongVideoBench\\(val)} & \makecell{MLVU\\(m-avg)} & \makecell{LVBench\\(avg)} & \makecell{TempComp\\(mc)} & \makecell{NExT-QA\\(test)} \\
        \midrule
        \multicolumn{10}{c}{\textit{1B Model Comparison}}\\
        \midrule
        SF-LLaVA-1.5-E2E-1B & \ding{52} & \textcolor{red}{\ding{56}} & 54.1 & 58.6 & 51.5 & 61.7 & \textbf{40.2} & 59.3 & 73.9 \\
        \rowcolor{lightblue}
        SF-LLaVA-1.5-1B & \textcolor{red}{\ding{56}} & \textcolor{red}{\ding{56}} & \textbf{56.6} & \textbf{61.9} & \textbf{54.3} & \textbf{64.3} & 39.7 & \textbf{60.5} & \textbf{76.7} \\
        \midrule
        \multicolumn{10}{c}{\textit{3B Model Comparison}}\\
        \midrule
        SF-LLaVA-1.5-E2E-3B & \ding{52} & \textcolor{red}{\ding{56}} & 58.4 & 62.4 & 53.0 & 65.0 & 40.9 & 63.2 & 78.6 \\
        \rowcolor{lightblue}
        SF-LLaVA-1.5-3B & \textcolor{red}{\ding{56}} & \textcolor{red}{\ding{56}} & \textbf{60.8} & \textbf{65.8} & \textbf{57.3} & \textbf{68.8} & \textbf{43.3} & \textbf{64.0} & \textbf{80.8} \\
        \midrule
        \multicolumn{10}{c}{\textit{7B Model Comparison}}\\
        \midrule
        SF-LLaVA-1.5-E2E-7B & \ding{52} & \textcolor{red}{\ding{56}} & 59.2 & 68.1 & 59.2 & 70.3 & 44.3 & 67.9 & 81.0 \\
        \rowcolor{lightblue}
        SF-LLaVA-1.5-7B & \textcolor{red}{\ding{56}} & \textcolor{red}{\ding{56}} & \textbf{63.9} & \textbf{69.6} & \textbf{62.5} & \textbf{71.5} & \textbf{45.3} & \textbf{68.8} & \textbf{83.1} \\
        \bottomrule
    \end{tabular}
    }
    \vspace{-5pt}
    \caption{
        \textbf{Results of SF-LLaVA-1.5-E2E on video benchmarks},
        which fully trains the visual encoder together with the projector and LLM.
    }
    \label{tab:e2e_video_results}
    \vspace{5pt}
\end{table}
\endgroup

\clearpage

\section{Qualitative Results}
\label{sec:qualitative_results}
\vspace{-10pt}

\begin{figure*}[h!]
    \centering
    \includegraphics[width=0.95\textwidth]{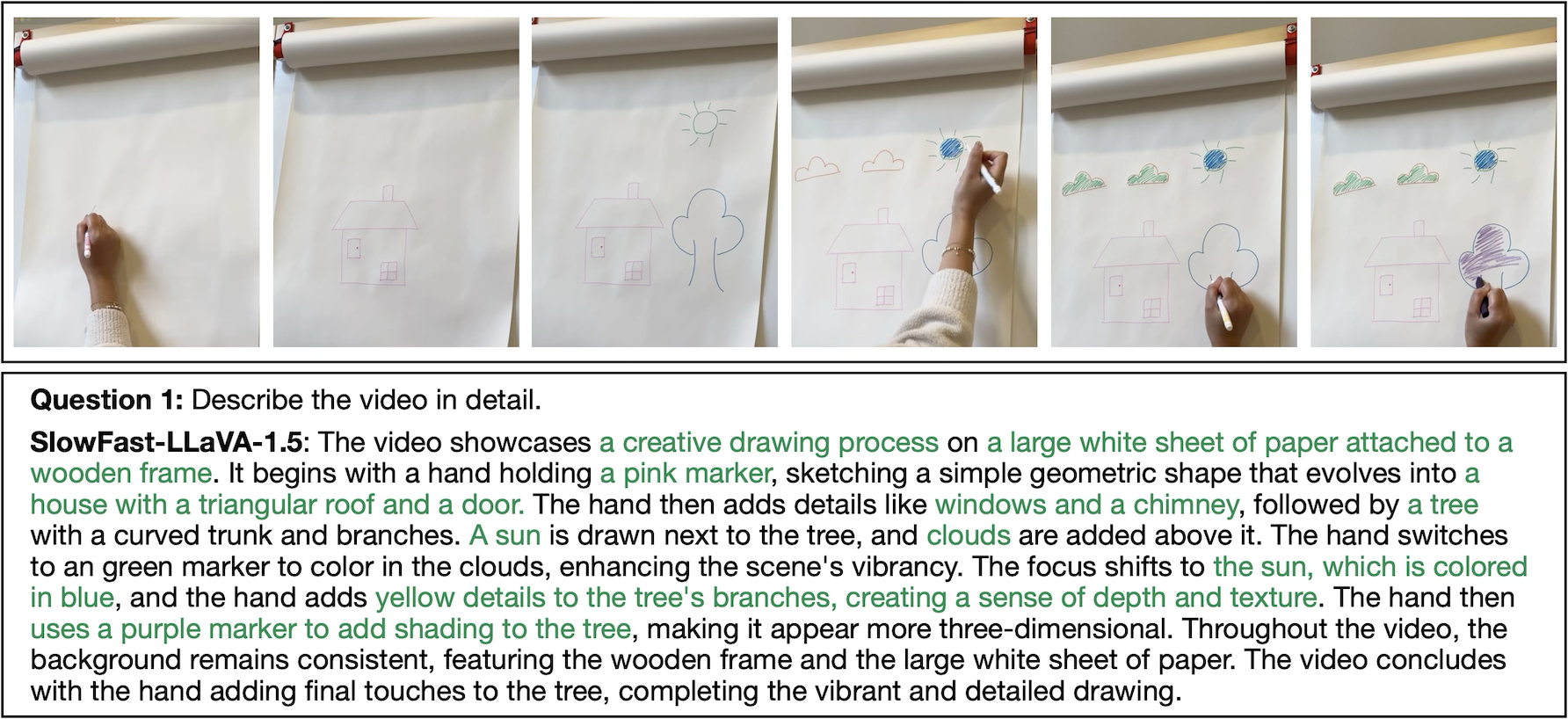}
    \vspace{-10pt}
    \caption{
        \textbf{SF-LLaVA-1.5 summarizes a video with detailed caption.}
    }
    \vspace{-5pt}
    \label{fig:sample_1}
\end{figure*}

\begin{figure*}[h!]
    \centering
    \includegraphics[width=0.95\textwidth]{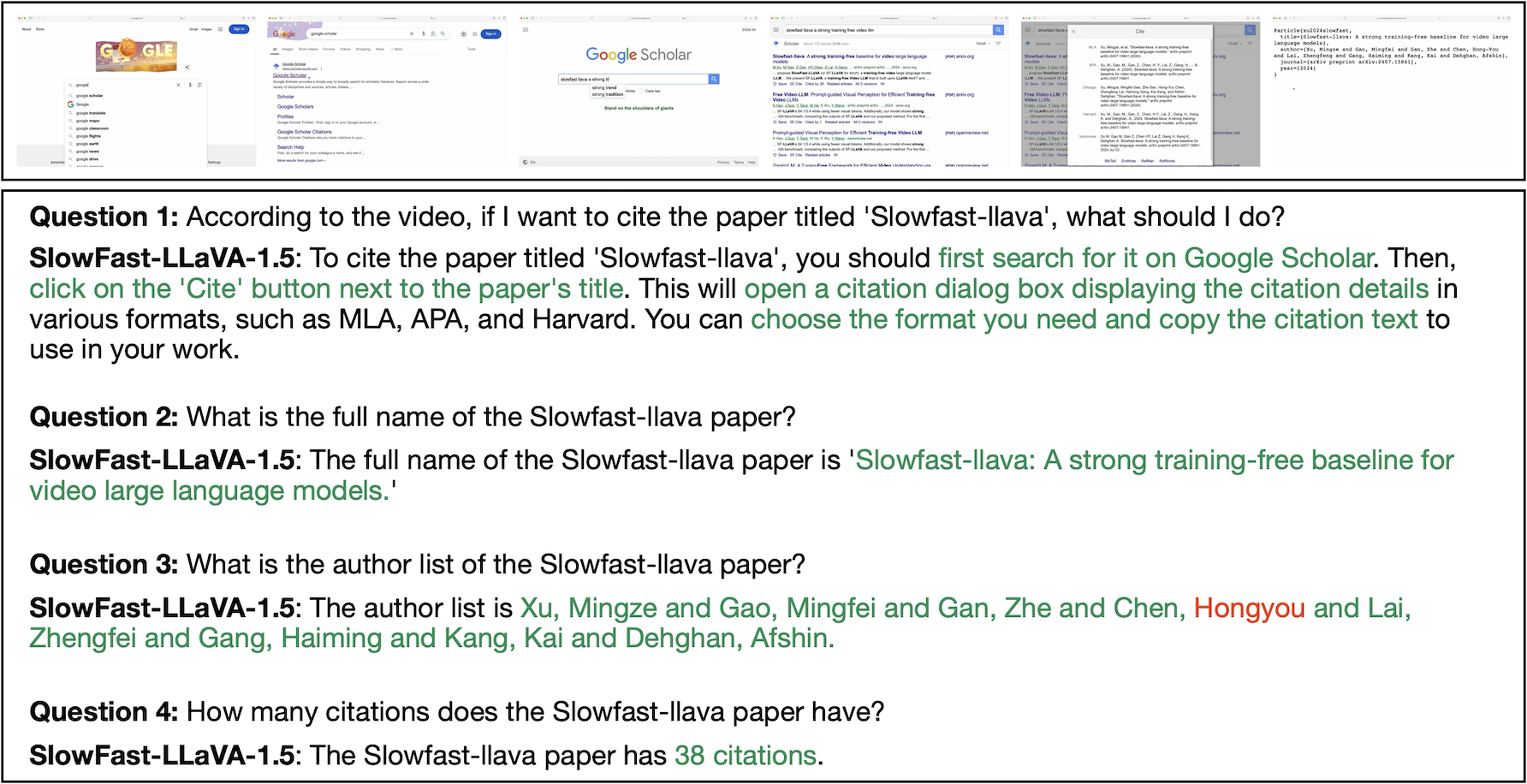}
    \vspace{-10pt}
    \caption{
        \textbf{SF-LLaVA-1.5 learns the process from the video and captures text-rich details.}
    }
    \vspace{-5pt}
    \label{fig:sample_2}
\end{figure*}

\begin{figure*}[h!]
    \centering
    \includegraphics[width=0.95\textwidth]{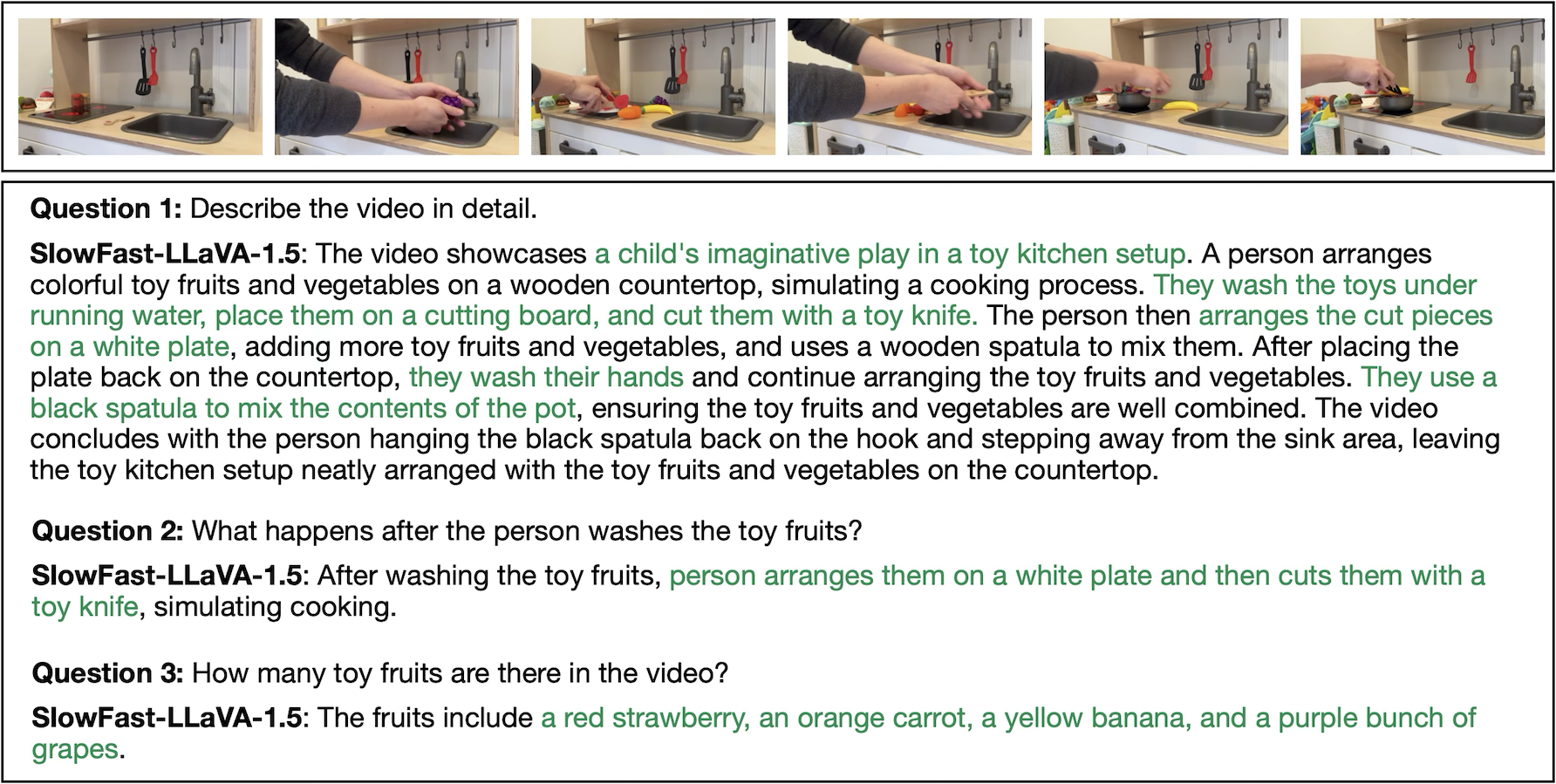}
    \vspace{-10pt}
    \caption{
        \textbf{SF-LLaVA-1.5 understands the relative sequence of different activities.}
    }
    \vspace{-5pt}
    \label{fig:sample_3}
\end{figure*}

\end{document}